\documentclass{article}
\usepackage{amsmath}
\usepackage{amsfonts}
\usepackage{graphicx} 
\usepackage{subcaption} 
\usepackage[margin=3cm]{geometry}
\usepackage{caption}
\usepackage[labelfont=bf, justification=raggedright, singlelinecheck=false]{caption}
\usepackage{float} 
\usepackage[table]{xcolor}
\definecolor{proposedcolor}{rgb}{0.85,1,0.85}  
\definecolor{bestcolor}{rgb}{0.85,0.92,1}       
\definecolor{improvecolor}{rgb}{0.85,1,0.85}  
\definecolor{regresscolor}{rgb}{1,0.85,0.85} 
\definecolor{ttacolor}{rgb}{0.92,0.95,1}  
\definecolor{revisioncolor}{rgb}{0.55,0.0,0.55}  
\date{}

\title{ThyCLIPNet: A BiomedCLIP-Guided Lightweight Attention-Enhanced DeepLabV3+ Framework for Robust Thyroid Nodule Segmentation}

\author{
  Tasnim Jahan$^{1}$,
  Md Easin Arafat$^{2,*}$,
  Swakkhar Shatabda$^{3,*}$ \\[2pt]
  \small $^{1}$Department of Computer Science and Engineering, United International University, \\
  \small United City, Madani Ave, Dhaka 1212, Bangladesh \\
  \small $^{2}$Faculty of Informatics, Institute of Industry-Academia Innovation, \\
  \small Department of Data Science and Engineering, Eötvös Loránd University, \\
  \small Pázmány Péter Sétány 1/C, Budapest 1117, Hungary \\
  \small $^{3}$Department of Computer Science and Engineering, BRAC University, \\
  \small Pragati Sarani, Merul Badda, Dhaka 1212, Bangladesh \\[4pt]
  \small $^{1}$\texttt{tasnimjahanjeny@gmail.com}, 
  $^{2}$\texttt{arafatmdeasin@inf.elte.hu}, 
  $^{3}$\texttt{swakkhar.shatabda@bracu.ac.bd} \\
  \small $^{*}$Corresponding authors
}

\usepackage[section]{placeins}
\usepackage{graphicx}
\usepackage{multirow}
\usepackage[colorlinks=true, linkcolor=blue, citecolor=blue, urlcolor=blue]{hyperref}

\begin{document}

\maketitle

\begin{abstract}
Accurate thyroid ultrasound segmentation is often challenged by low contrast, speckle noise, and unclear boundaries. Although recent methods have improved segmentation accuracy, many rely on resource-intensive architectures or lack explicit integration of multiscale features with global biomedical visual guidance. In this paper, we introduce ThyCLIPNet, a novel lightweight semantic-guided hybrid encoder-decoder framework that integrates BiomedCLIP-derived biomedical semantic guidance into a lightweight multi-scale CNN segmentation pipeline while maintaining low computational overhead. The encoder integrates MobileNetV2 with efficient channel attention to refine hierarchical multi-scale features, while an adapted atrous spatial pyramid pooling module combines with a custom convolutional block attention module at the bottleneck to enrich semantic context and refine features. The decoder combines hierarchical skip connections and lightweight attention refinement with a BiomedCLIP-guided gated fusion pathway that projects vision-only global biomedical embeddings into decoder feature space and selectively integrates them at lower-resolution stages through semantic-local fusion and spatial gating. To the best of our knowledge, ThyCLIPNet is among the first lightweight thyroid ultrasound segmentation frameworks to employ BiomedCLIP’s vision encoder alone for image-only global semantic guidance without text prompting. This image-only semantic-conditioning strategy is particularly suited to thyroid ultrasound datasets that provide images and segmentation masks without paired clinical text. Extensive experiments on four publicly available thyroid ultrasound datasets, TG3K, TN3K, DDTI, and PKTN, demonstrate the effectiveness of ThyCLIPNet. The model achieves dice similarity coefficients of 96.22\%, 87.58\%, 84.73\%, and 80.70\%; intersection over union scores of 92.72\%, 77.91\%, 73.51\%, and 67.64\%; and 95th‑percentile hausdorff distances of 3.75, 16.38, 18.23, and 10.86, respectively. These results demonstrate competitive segmentation performance across the evaluated datasets while maintaining a compact segmentation-stage configuration of 8.55 M parameters and 22.99 G FLOPs. Overall, the findings support selectively integrating global biomedical semantic guidance with lightweight multi-scale CNN representations as an effective strategy for robust and computationally efficient thyroid ultrasound segmentation. The source code for our proposed model is publicly available at: \url{https://github.com/Tasnim-Jahan/ThyCLIPNet}
\end{abstract}


\section{Introduction}
The thyroid gland is a butterfly-shaped endocrine gland that plays a critical role in maintaining normal physiological functions in the human body \cite{al2022thyroid, chang2010thyroid}. Among various thyroid disorders, thyroid nodules are particularly common, typically forming within the thyroid gland \cite{alexander2022diagnosis}. The American College of Radiology Thyroid Imaging Reporting and Data System (TI-RADS) \cite{tessler2017acr} reveals that the morphology and borders of thyroid nodules are important features for the diagnosis of malignant nodules. Nowadays, thyroid diseases have become increasingly common, with an annual incidence rate exceeding 10\% in the last decade \cite{chi2023hybrid}. Due to the potential for malignancy, regular monitoring of the thyroid is crucial to enable early detection, accurate diagnosis, and timely intervention, thereby improving patient outcomes and reducing the risk of disease progression \cite{durante2018diagnosis}.

Thyroid nodules are commonly evaluated using fine-needle aspiration (FNA), computed tomography (CT), magnetic resonance imaging (MRI), and ultrasound (US) \cite{nie2022n}. Although CT and MRI typically yield only incidental findings and are not recommended as primary diagnostic tools for thyroid nodules \cite{boers2023ultrasound}, FNA is widely considered the gold standard for confirming malignancy, despite being traumatic to the human body \cite{baloch2002diagnosis}. Ultrasound is a non-invasive, radiation-free method that is cost-effective and quick, capable of detecting lesions as small as a few millimetres. Owing to these advantages, it is widely employed for thyroid evaluation \cite{wang2020automatic}. However, ultrasound diagnosis relies heavily on physician expertise, which can lead to misdiagnosis. This is where computer-aided diagnosis (CAD) systems can play a significant role, as they have the potential to enhance diagnostic accuracy and assist clinicians in their decision-making processes \cite{li2018fully, park2019diagnosis}.

Among various CAD approaches, deep learning-based methods are widely used to automatically trace thyroid nodules and improve diagnostic consistency. Attention-based methods, such as Swin U-Net \cite{yang2025improving}, have improved edge preservation and spatial feature extraction through self-attention mechanisms. In contrast, an AttU-Net \cite{cozzi2025multi} architecture developed for thyroid nodule segmentation demonstrated robust generalization when evaluated on data from multiple clinical centers. Transformer-based method, such as Enhanced-TransUNet \cite{ozcan2024enhanced}, combines transformer encoding with UNet decoders and an information bottleneck to mitigate overfitting while preserving contextual features. Furthermore, the improved Swin Transformer incorporates depthwise convolutions and multi-scale fusion to enhance feature quality in noisy environments \cite{wu2025thyroid}. Diffusion-based models like PEW-SegDiff \cite{bi2025pew} used feature pyramids and edge-weighted attention to enhance boundary delineation under low-contrast conditions. Beyond segmentation, other frameworks have been proposed that jointly perform multiple tasks; for instance, FADLM \cite{zhang2025automatic} integrated Mask R-CNN for automatic thyroid primary tumor segmentation and ResNet34 with a Bayes strategy for cervical LNM diagnosis, while TNVis \cite{zhou2025deep} used dynamic ultrasound keyframes and 3D visualization to align more closely with clinical workflows. 

In addition, MAUNet \cite{xiang2025federated} applied federated learning with multi-scale and dual attention to achieve strong generalization across multi-center datasets without compromising data privacy. Multi-view learning frameworks such as MLMSeg \cite{chen2024mlmseg} captured local, global, and structural feature correlations through cross-layer graph fusion, improving robustness across imaging variations. Weakly supervised strategies have gained attention for reducing reliance on dense annotations. Additionally, a multi-branch model that integrates spatial, contextual, and multi-scale consistency has further enhanced segmentation accuracy using bounding box supervision \cite{aly2025weakly}. Despite these advancements, many existing approaches either prioritize segmentation accuracy at the cost of efficiency or rely on complex architectures that are impractical for broader clinical integration. These limitations highlight the need for a lightweight yet effective segmentation framework that can preserve detailed local representations while incorporating broader semantic context for thyroid ultrasound images. Robust delineation of thyroid structures is therefore important for reliable ultrasound-based assessment and subsequent clinical decision-making.

In this paper, we introduce ThyCLIPNet, a lightweight semantically guided framework for thyroid ultrasound segmentation that integrates multiscale CNN representation learning with global biomedical visual guidance. The architecture adopts a DeepLabV3+-style encoder and U-Net-style decoding pathway, with BiomedCLIP-derived global visual representations selectively introduced into the decoder through the proposed BiomedCLIP-guided Gated Fusion (BGF) module. Unlike existing CLIP-assisted thyroid ultrasound segmentation approaches, ThyCLIPNet employs BiomedCLIP's frozen vision encoder alone as a global biomedical semantic prior, without text prompting, text-encoder involvement, or dense token-level interaction, thereby preserving computational efficiency while enriching semantic context for challenging ultrasound images. The key contributions of this work are as follows:

\begin{itemize}

\item Introduced a BiomedCLIP-guided Gated Fusion (BGF) module that uses the frozen BiomedCLIP vision encoder as a source of global biomedical semantic priors and selectively conditions low-resolution decoder representations through semantic-local fusion, spatial gating, and feature concatenation, without relying on text prompting or multimodal supervision, thereby preserving lightweight inference.

\item Enhanced the encoder with a MobileNetV2 backbone by incorporating an Efficient Channel Attention (ECA) module and a bridge bottleneck that combines an adapted Atrous Spatial Pyramid Pooling (ASPP) module with a custom Convolutional Block Attention Module (CBAM), enriching semantic representations without additional downsampling or computational burden.

\item Designed a three-stage decoder with skip connections, lightweight CBAM blocks, and selective BiomedCLIP-guided semantic conditioning to progressively refine spatial accuracy and boundary delineation.

\item Validated the proposed framework on four benchmark thyroid ultrasound datasets (TG3K, TN3K, DDTI, and PKTN), demonstrating competitive segmentation performance together with strong computational efficiency in terms of model size, training time, and inference latency.
\end{itemize}

The remainder of this paper is organized as follows.
\hyperref[sec:literature]{Section~\ref*{sec:literature}} provides a comprehensive review of the related literature. \hyperref[sec:dataset]{Section~\ref*{sec:dataset}} describes the datasets, experimental setup, and overall methodology, followed by a detailed explanation of the proposed ThyCLIPNet architecture. \hyperref[sec:results]{Section~\ref*{sec:results}} presents the experimental results along with a thorough discussion and comparative analysis.
Finally, \hyperref[sec:conclusion]{Section~\ref*{sec:conclusion}} concludes the paper and outlines directions for future work.

\section{Related Works}
\label{sec:literature}
Recent studies in Biomedical Image Segmentation (BIS), especially in thyroid ultrasound, have utilized deep learning architectures to enhance segmentation performance. In this section, we will review the evolution from traditional handcrafted methods to more advanced techniques. These include Convolutional Neural Networks (CNNs), attention-guided networks, transformers, hybrid lightweight designs, and emerging CLIP-based vision-language models, particularly those that have been validated on benchmark thyroid ultrasound datasets.

\subsection{Traditional Thyroid Ultrasound Segmentation}
Early thyroid ultrasound segmentation techniques primarily relied on classical computer vision methods, including region-based approaches \cite{295913}, threshold-based methods \cite{otsu1975threshold}, and deformable model-based techniques, which were originally introduced through active contour models \cite{kass1988snakes}  and later extended to medical image analysis using a level set approach \cite{malladi2002shape}. Region-based segmentation expands or reduces areas based on similarity rules to group similar pixels \cite{295913}, while threshold-based segmentation classifies pixels based on intensity values derived from grayscale distributions \cite{otsu1975threshold}. Deformable models segment shapes by fitting known geometries, such as spheres and cones, to point cloud data, using features like brightness and texture, and are efficient and noise resilient \cite{kass1988snakes}. However, these approaches rely on manual feature engineering, and limited adaptability to complex anatomical variations hinders generalizability. 

\subsection{CNN-Based Segmentation Architectures} 
Convolutional Neural Networks (CNNs) have revolutionized medical image segmentation, beginning with the pioneering Fully Convolutional Network (FCN) \cite{long2015fully}. This model replaced the need for handcrafted feature design with end-to-end learning, paving the way for U-Net \cite{ronneberger2015u}, a foundational architecture that continues to inspire numerous specialized CNN-based models in the field of medical imaging. Inan et al. \cite{inan2024multi} proposed a hybrid ResUNet and ResUNet++ framework that combines residual blocks with the encoder–decoder structure of U-Net, integrating efficient feature extraction and upsampling for multi-class thyroid nodule segmentation. Bouhdiba et al. \cite{bouhdiba2024ultrasound} improved upon the traditional U-Net by incorporating residual blocks in its encoder–decoder pathway, improving the model's capacity to handle complex thyroid textures under noise and low contrast. Similarly, Peng et al. \cite{peng2024enhanced} introduced the DC-Contrast U-Net, a lightweight CNN-based model specifically designed for pediatric thyroid segmentation. This model focuses on efficient feature encoding while minimizing overall complexity. Xing et al. \cite{xing2024real} developed FCDDN, a fully convolutional architecture that incorporates dense connectivity, dilated convolution, and factorized filters to maintain high segmentation efficiency with minimal parameters. Furthermore, EfficientNet-B7 has been adopted as the backbone for DeepLabV3+, demonstrating high segmentation accuracy when faced with common ultrasound challenges such as noise and blurred edges. Recently, Gummalla et al. \cite{gummalla2024enhanced} developed a sequential CNN paired with K-means clustering for extracting regions of interest, highlighting CNNs' utility even in hybrid detection-classification pipelines. Additionally, Etehadtavakol et al. \cite{etehadtavakol2024enhanced} introduced a fusion architecture that combines U-Net and VGG16, enhanced with feature engineering and radiomics, to improve the delineation of thyroid nodule boundaries, particularly in thermal images. This underscores the effectiveness of using efficient and performance-preserving backbone architectures for thyroid imaging tasks \cite{xiao2025ultrasound}. Collectively, these models demonstrate the flexibility and segmentation capability of CNN-based approaches across various thyroid imaging scenarios.

\subsection{Attention-Driven Architectures}
Specialized attention mechanisms, such as self-attention, dual-path attention, and multiscale attention, have significantly enhanced thyroid ultrasound segmentation by effectively addressing noise, low resolution, and boundary ambiguity. For instance, Wang et al. \cite{wang2024via} presented this approach through the integration of multi-scale cross attention and dual attention within skip connections, enhancing encoder–decoder interaction to better capture variations in nodule size and structure. Similarly, DPAM-UNet++ \cite{dong2024ultrasound} incorporates a dual-path attention mechanism that enables the model to capture both global and local contextual information, significantly improving edge precision. Further, MLCA-UNet \cite{zhou2025mlca} combines multiscale linear attention with convolutional attention, effectively balancing global feature diversity and computational efficiency. In addition, an enhanced Swin U-Net \cite{yang2025improving} leverages self-attention mechanisms, residual connections, and multiscale convolutional modules embedded within long skip connections, thereby enabling robust edge preservation and effective feature extraction. Furthermore, Li et al. \cite{li2025multi} proposed a multi-scale dynamic sparse attention module that dynamically filters and refines fine-grained cross-scale interactions, while enhanced downsampling and upsampling modules within a U-Net–like architecture contribute to achieving high segmentation accuracy with reduced model complexity on thyroid datasets. Complementing these, CIL-Net \cite{ali2024cil} fuses triplet attention, dilated convolution, and residual context modeling to learn fine-grained context while effectively managing irregularities in nodule shape.

\subsection{Transformer-Based Architectures}  
Recently, transformer-based architectures have become integral to thyroid ultrasound segmentation due to their strong global context modeling and adaptability to complex anatomical variations. Deng et al. \cite{deng2024stu3net} proposed STU$^3$Net, a hybrid network combining CNN encoders with swin transformer modules for effective extraction of morphological features and edge details in thyroid ultrasound by embedding transformer features into a three-layer U-Net structure with cross-layer skip connections. In another study, TransUMobileNet \cite{cai2025transumobilenet} fused a hybrid CNN–Transformer encoder with a multi-channel attention fusion module. This combination enabled fine-grained boundary refinement and global semantic encoding, yielding high segmentation performance on a custom thyroid dataset. Beyyala et al. \cite{beyyala2024swin} employed the swin transformer and attention-guided segmentation model, combining transformer-based encoders with prior-guided modules and attention-enhanced decoders, demonstrating performance gains in the DDTI dataset. Finally, SAM-MedUS \cite{tian2025sam} was developed as a foundational model for general-purpose ultrasound segmentation. While it was not specifically designed for thyroid imaging, it included ConvNeXt V2 and channel attention modules to ensure robust global context extraction. Furthermore, it employed boundary loss to handle low-contrast and fuzzy anatomical regions.

\subsection{Hybrid and Lightweight Architectures} 
More recent studies on thyroid ultrasound segmentation have increasingly focused on hybrid and lightweight architectures. These designs incorporate efficient encoders such as MobileNetV2 \cite{DBLP:journals/corr/abs-1801-04381}, EfficientNet \cite{Tan2019EfficientNetRM}, ShuffleNetV2 \cite{Ma_2018_ECCV}, and GhostNet \cite{Han_2022}, which deliver strong representational capacity with minimal computational overhead. One notable development in this area is GSE-Nets \cite{li2025gse}, which integrates a global structure enhancement decoder with structure cross-attention, a lightweight graph convolution block, and feature refinement attention, improving structural representation and boundary clarity while combined with different encoders to construct a set of segmentation networks. Another development in this field is the multi-scale adaptive convolutional network with DBUNet \cite{zhu2025dual} (MSAC-DBUNet) and its lightweight variant, named MSAC-DD \cite{begum2025smart, payatsuporn2025papillary}, proposed for Papillary Thyroid Carcinoma (PTC) histology image segmentation, employs an auto-adjusting multi-scale kernel convolution encoder and a dual-decoder design incorporating the Feature Channel Interaction Attention (FCIA) module. Furthermore, GDSSA-Net \cite{farooq2025gdssa} enhances accuracy while remaining efficient for resource-constrained, real-time applications by incorporating gated attention in the decoder and employing a gradually deep supervision strategy, supported by a self-ensembling design.
\subsection{CLIP-based Architectures}
Vision-language pretraining has only recently been explored for thyroid ultrasound segmentation, introducing new opportunities to incorporate pretrained semantic representations into ultrasound segmentation frameworks. Most recently, Sun et al. proposed CLIP-TNseg \cite{sun2025clip}, a multimodal hybrid framework that combines textual and high-level visual semantic information from a frozen CLIP branch with fine-grained features extracted through a U-Net-style segmentation branch. Building on this line of work, Kang et al. proposed GCAESeg \cite{kang2025gcaeseg}, a CLIP-guided thyroid ultrasound segmentation framework that incorporates visual-linguistic semantic information with U-shaped visual representations through grouped attention mechanisms, improving segmentation particularly for boundary-ambiguous and small lesions on PKTN compared to CLIP-TNseg \cite{sun2025clip}. Beyond thyroid-specific designs, MedCLIP-SAMv2 \cite{KOLEILAT2025103749} integrates BiomedCLIP with the Segment Anything Model \cite{kirillov2023segment} using text-guided BiomedCLIP representations and an M2IB-based mechanism to generate visual prompts for zero-shot segmentation, while also investigating weakly supervised training across ultrasound, MRI, X-ray, and CT. Similarly, Grounding DINO-US-SAM \cite{11146904} combines Grounding DINO \cite{liu2024grounding} with SAM2-based mask generation in a prompt-driven vision-language framework, enabling multi-organ ultrasound segmentation, including thyroid, with strong performance on held-out unseen ultrasound datasets. Recent SAM2-based medical segmentation work, such as SAM2-UNet \cite{xiong2024sam2}, further reflects the growing interest in adapting foundation-model representations for medical imaging. In addition, adapter-based SAM approaches, such as Medical SAM Adapter (Med-SA) \cite{wu2025medical}, demonstrate that lightweight adapter mechanisms can enhance medical image segmentation by updating only a small fraction of parameters. These studies demonstrate the potential of pretrained vision-language representations, but they leave a different design question: how biomedical semantic guidance can be incorporated into a lightweight thyroid ultrasound segmentation pipeline without requiring textual prompting or multimodal interaction. Prior thyroid ultrasound segmentation studies have not specifically investigated BiomedCLIP’s vision encoder alone as a source of global biomedical semantic guidance selectively integrated with local decoder features in a lightweight image-only framework. This specific gap motivates ThyCLIPNet, which combines a lightweight CNN segmentation pipeline with selective BiomedCLIP-guided semantic conditioning to incorporate global biomedical semantic visual priors while preserving computational efficiency for thyroid ultrasound segmentation.


\section{Methods and Materials}
\label{sec:dataset}
This section outlines the datasets used in the study, describes the pre-processing strategies, and presents the architectural foundations along with details of the proposed ThyCLIPNet framework. It concludes with the experimental setup and evaluation methodology.


\subsection{Dataset Description}
We evaluated our proposed method using four publicly available thyroid ultrasound datasets, namely Thyroid Gland (TG3K) \cite{wunderling2017comparison}, Thyroid Nodule (TN3K) \cite{gong2023thyroid}, Digital Database Thyroid Image (DDTI) \cite{pedraza2015open}, and Peking University Thyroid Nodule (PKTN) \cite{sun2025clip}. These datasets encompass two related but distinct thyroid ultrasound segmentation tasks: TG3K evaluates thyroid gland segmentation, whereas TN3K, DDTI, and PKTN evaluate thyroid nodule segmentation. The TG3K dataset comprises 3,585 curated frames extracted from 16 ultrasound videos, with frames retained when the thyroid gland occupies more than 6\% of the image area to form a 2D thyroid gland segmentation benchmark. The TN3K dataset comprises 3,493 grayscale ultrasound images from 2,421 patients, each image featuring at least one thyroid nodule, no blood signal, and a single representative view per patient location. The images were cropped for consistency, and labels were provided by volunteers under the guidance of a radiologist, with confirmation by both a radiologist and a senior physician. Each subset of the TN3K dataset contains images taken from a single patient, ensuring independence between the training set (2,879 images) and the test set (614 images). The DDTI dataset includes 637 images with pixel-level lesion masks, offering varying resolutions and representative lesion structures. Finally, the PKTN dataset consists of 1,005 images from patients at Peking University First Hospital, featuring both benign and malignant nodules with expert-reviewed annotations to ensure precise boundary delineation, especially in diagnostically challenging cases. These datasets provide a robust and diverse benchmark for evaluating segmentation performance across multiple thyroid ultrasound imaging scenarios. 

\hyperref[tab:dataset_info]{Table~\ref*{tab:dataset_info}}  summarizes the key characteristics and split sizes of the four thyroid ultrasound datasets used in our evaluation. We report image counts before and after augmentation to keep the pipeline transparent. We split TG3K, DDTI, and PKTN into training, validation, and test sets, while for TN3K, we retain the official test set and split only the remaining data into training and validation. The \textit{Total images (original)} column denotes the number of unique images available in each dataset prior to augmentation. The \textit{Train (original)}, \textit{Validation}, and \textit{Test} columns report the exact counts of samples in each split before augmentation. Offline augmentation is applied exclusively to the training set. The augmentation factor ($k$) represents the number of synthetic variants generated per original training image. Accordingly, the number of augmented training images is $k \times$ (train images, original), the final training set size becomes $(k+1) \times$ (train images, original), and the \textit{Total images (after augmentation)} equals the sum of \textit{Total (original)} and the augmented training images.

 \begin{table*}[ht]
\caption{Key characteristics and dataset splits for model evaluation.}
\label{tab:dataset_info}
\centering
\resizebox{\textwidth}{!}{%
\begin{tabular}{l c c c c c c c c c}
\hline
\textbf{Dataset} & \textbf{\shortstack{Total\\Images\\(Original)}} & \textbf{\shortstack{Train\\Images\\(Original)}} & \textbf{\shortstack{Validation\\Images}} & \textbf{\shortstack{Test\\Images}} & \textbf{\shortstack{Augmentation\\Factor (k)}} & \textbf{\shortstack{Augmented\\Train\\Images}} & \textbf{\shortstack{Final\\Train\\Images}} & \textbf{\shortstack{Total\\Images\\(After Augmentation)}} & \textbf{\shortstack{Ultrasound\\Devices}} \\
\hline
TG3K \cite{wunderling2017comparison}  & 3,585 & 2,868 & 358 & 359 & 1 & 2,868 & 5,736 & 6,453 & GE Logiq E9 \\
TN3K \cite{gong2023thyroid}  & 3,493 & 2,304 & 575 & 614 & 1 & 2,304 & 4,608 & 5,797 & GE Logiq E9; ARIETTA 850; RESONA 70B \\
DDTI \cite{pedraza2015open} & 637   & 509   & 63  & 65  & 5 & 2,545 & 3,054 & 3,182 & Toshiba Nemio 30; Toshiba Nemio MX \\
PKTN \cite{sun2025clip} & 1,005 & 804   & 100 & 101 & 4 & 3,216 & 4,020 & 4,221 & Not disclosed \\
\hline
\end{tabular}%
}
\end{table*}

\subsection{Image Preprocessing and Augmentation}
\begin{figure*}[htbp]
  \centering
  \includegraphics[width=\textwidth]{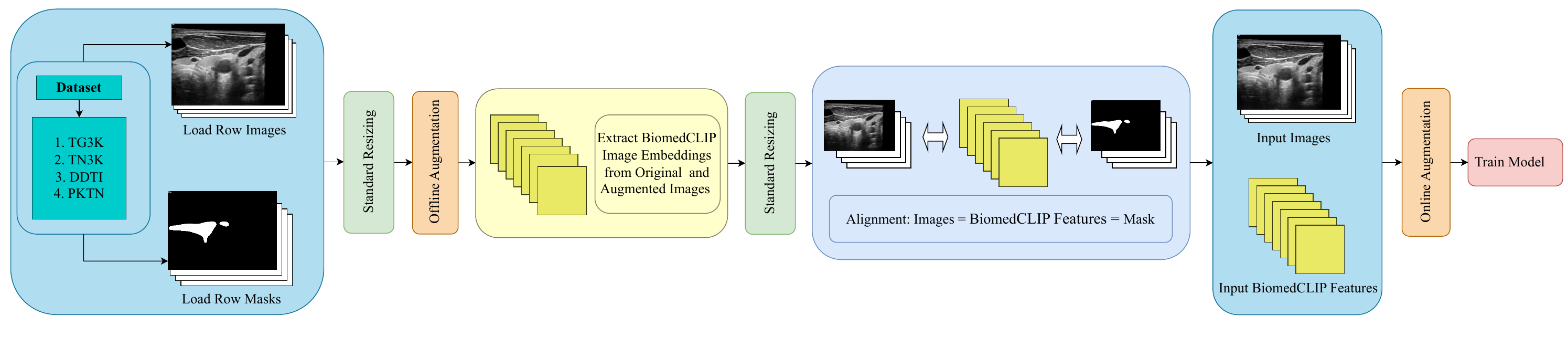}
  \captionsetup{justification=justified}
\caption{Illustration of the preprocessing and data‐augmentation pipeline: ultrasound images and masks are resized to 256$\times$256 and normalized; dataset-specific offline augmentations are applied; 512-dimensional BiomedCLIP embeddings are extracted from original and augmented images; dynamic online augmentations are performed on both images and embeddings; and the resulting data are fed into the segmentation model for training.}
  \label{fig:data_pipeline}
\end{figure*}

To enhance the robustness and generalizability of our method across the TG3K, TN3K, DDTI, and PKTN datasets, we resized all ultrasound images and their corresponding masks to a uniform resolution of 256$\times$256 pixels. We then normalized pixel intensities to ensure consistency across varying acquisition settings. To address data scarcity and class imbalance, we employed a two-stage augmentation strategy. In the offline stage, we applied dataset-specific augmentations using the albumentations library, with augmentation factors scaled based on the size of each dataset. These included spatial transformations (horizontal flips, affine translations, rotations, and scaling), elastic deformations, and intensity modifications (brightness, contrast, and gamma). Additional variations, such as Gaussian noise, blurring, coarse dropout, and shadow simulation, were used to expose the model to diverse imaging artifacts. Following this, we extracted 512-dimensional image embeddings from both the original and augmented images using BiomedCLIP, employing only the visual modality without textual input in training or inference. These embeddings provided rich semantic representations that enhanced anatomical feature recognition. During training, we dynamically applied online augmentations to both the images and their corresponding BiomedCLIP-based embeddings, further improving model generalization without increasing storage requirements. For unbiased evaluation, we kept the validation and test sets unaugmented. \hyperref[fig:data_pipeline]{Figure~\ref*{fig:data_pipeline}} illustrates the complete preprocessing and training pipeline, from dataset preparation to final model training.

\subsection{Overview of DeeplabV3+}
DeepLabV3+ \cite{chen2018encoder} extends the atrous‐only DeepLabV3 by adding a simple yet effective decoder module to sharpen object boundaries. The encoder utilizes a deep backbone, typically Xception or ResNet-101, which is pre-trained on ImageNet \cite{5206848} to extract multi-scale features. The Atrous Spatial Pyramid Pooling (ASPP) captures multi-scale context by applying one 1$\times$1 convolution and three 3$\times$3 convolutions (dilation 6, 12, and 18 for output stride 16) in parallel. It then aggregates the outputs with a global pooled feature through concatenation and a 1$\times$1 convolution, which helps preserve resolution and enrich semantic information. The decoder restores spatial details by projecting low-level features from the early layers of the encoder backbone into fewer channels. It concatenates these features with the upsampled ASPP output and refines the result through two 3$\times$3 convolutions, followed by a final 4$\times$ upsampling to match the original image size. DeepLabV3+ has achieved state-of-the-art performance on the PASCAL VOC 2012 and Cityscapes benchmarks by effectively balancing boundary delineation with global context aggregation. However, a limitation of DeepLabV3+ is its dependence on parameter-intensive backbones, which can lead to increased model size and slower inference. To address this, we replace the backbone with MobileNetV2 \cite{sandler2018mobilenetv2}, a lightweight yet powerful encoder, enrich the bottleneck with Efficient Channel Attention (ECA) \cite{wang2020eca} and Convolutional Block Attention Module (CBAM) variants \cite{woo2018cbam}, along with BiomedCLIP-derived global image embeddings \cite{zhang2024biomedclip} to improve focus on fine-grained ultrasound imagery. 

\subsection{Proposed Architecture}
The proposed segmentation model introduces a hybrid deep learning architecture that integrates DeepLabV3+’s multi-scale context encoding, UNet-style skip-based decoder refinement, and BiomedCLIP-guided semantic fusion to enhance ultrasound image segmentation. The encoder employs MobileNetV2 as a lightweight and effective backbone for extracting hierarchical features from the input image. These feature maps are subsequently refined using ECA modules, which strengthen channel-wise dependencies without increasing spatial complexity. The output from the deepest encoder layer is processed through an ASPP module to capture diverse receptive fields and integrate multi-scale features. To further strengthen the bottleneck representation, we incorporate a custom CBAM module that performs spatial and channel-wise refinement. This increases the model's capacity to focus on diagnostically relevant structures.  

\begin{figure*}[!ht]
  \centering  \includegraphics[width=\textwidth,keepaspectratio]{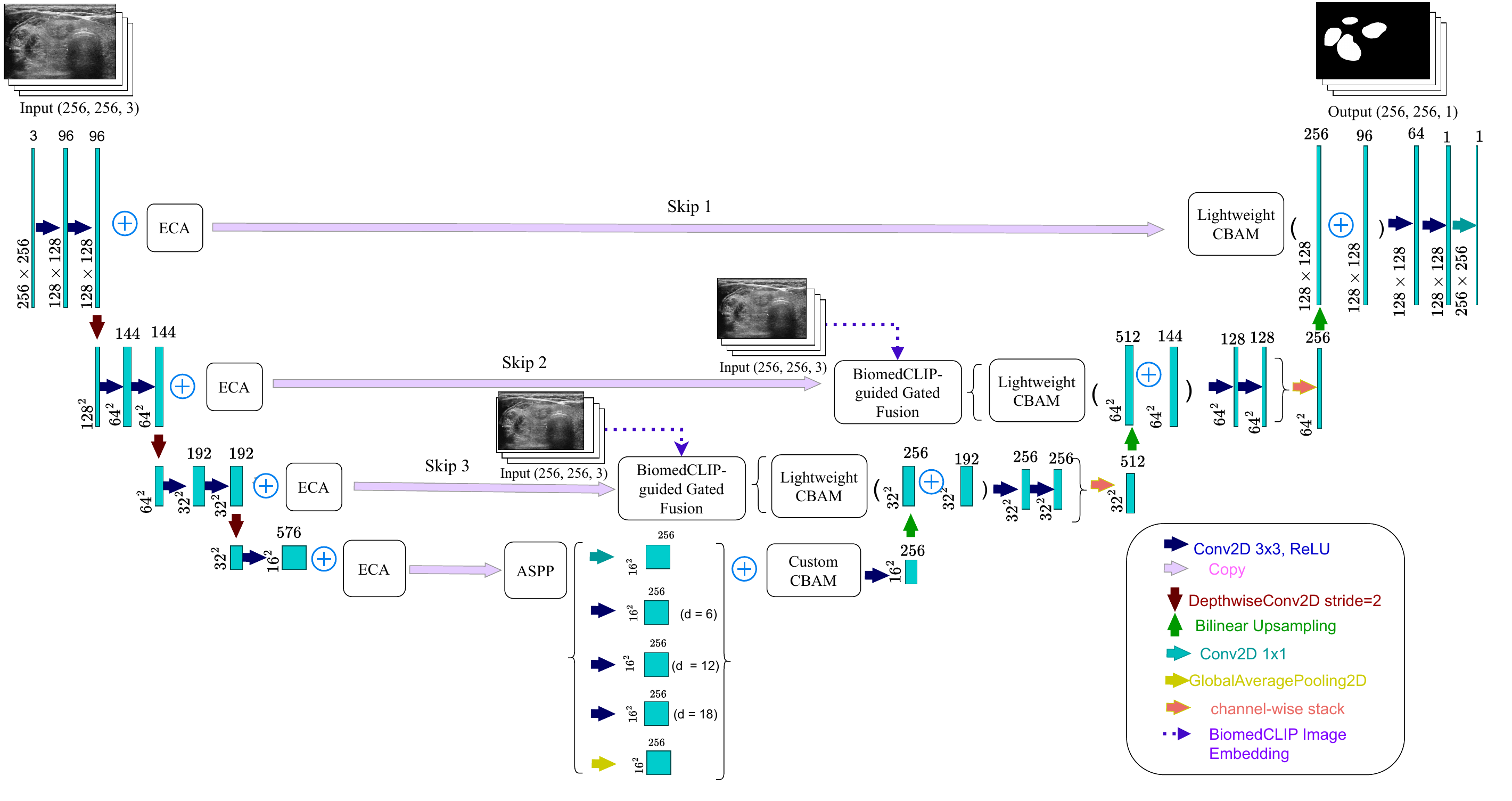}
  \captionsetup{justification=justified}
  \caption {Architecture of the proposed ThyCLIPNet framework for thyroid ultrasound segmentation, illustrating: (a) MobileNetV2 encoder integrated with ECA modules; (b) multi-scale context capture using an ASPP block fused with a custom CBAM in the bottleneck; (c) global semantic guidance from BiomedCLIP image embeddings integrated through the BiomedCLIP-Guided Gated Fusion (BGF); and (d) three-stage U-Net–style decoder with skip connections and lightweight CBAM for progressive boundary refinement.}
  \label{fig:proposed-f}
\end{figure*}

A key contribution of the architecture is the integration of BiomedCLIP-derived global image embeddings obtained solely from its vision encoder. We employ BioMedCLIP because it is pretrained on biomedical image–text pairs, providing medically grounded semantic priors that are more domain-relevant than generic vision backbones. Importantly, our objective is not multimodal grounding but lightweight semantic regularization: the BiomedCLIP branch injects global anatomical context that helps disambiguate low-contrast thyroid boundaries under speckle noise. To preserve real-time feasibility, we restrict the BiomedCLIP pathway to a single global embedding rather than dense token-level fusion. These global features are integrated into the deeper decoding stages, where they are fused with upsampled decoder outputs while skip connections and lightweight CBAM modules progressively refine spatial details. The final output layer generates a binary thyroid mask, followed by bilinear upsampling to match the original input resolution. \hyperref[fig:proposed-f]{Figure~\ref*{fig:proposed-f}} provides an overview of the proposed architecture.

\subsubsection{Encoder Pathway}
The encoder pathway transforms the input ultrasound image into progressively lower resolution, higher-semantic feature maps that guide the decoder through skip connections. To achieve a balance between computational efficiency and semantic depth, we have chosen MobileNetV2 as the encoder backbone. MobileNetV2 employs depthwise separable convolutions and inverted residual blocks, providing an optimal trade-off between speed and accuracy, particularly suited for low-contrast grayscale medical images.
As the input image flows through the encoder, it is progressively transformed into a hierarchy of feature representations at different depths. To capture this multi-scale contextual information, we select four representative feature maps from the hierarchy, specifically from block 1 \texttt{expand\_relu}, block 3 \texttt{expand\_relu}, block 6 \texttt{expand\_relu}, and block 13 \texttt{expand\_relu}. These selected layers produce feature maps corresponding to progressively deeper stages of the network, denoted as $F_1, F_2, F_3,$ and $F_4$, respectively. For an input of size $256 \times 256$, the dimensionality and representational roles of these feature maps are as follows:

\[
\begin{aligned}
F_1 &\in \mathbb{R}^{128 \times 128 \times 96},  && \text{capturing fine spatial details, such as edges and contours}, \\
F_2 &\in \mathbb{R}^{64 \times 64 \times 144}, && \text{encoding intermediate textures and structures}, \\
F_3 &\in \mathbb{R}^{32 \times 32 \times 192}, && \text{representing higher-order semantic abstractions}, \\
F_4 &\in \mathbb{R}^{16 \times 16 \times 576}, && \text{containing deep contextual representations}.
\end{aligned}
\]

These multiscale features serve as skip connections and are later fused into the decoder. Although they capture information across a wide range of spatial resolutions and semantic levels, the segmentation performance can be further improved by enhancing interactions between channels. This is particularly relevant in ultrasound imaging, where subtle grayscale variations often convey critical diagnostic information. To address this, we employ an ECA \cite{wang2020eca} module uniformly across all encoder outputs to adaptively recalibrate channel responses.
Given a feature map \( F \in \mathbb{R}^{H \times W \times C} \), where \( C \) denotes the number of channels. We compute a global channel descriptor \( z \in \mathbb{R}^C \) by spatially averaging each channel:
\begin{equation}
z_c = \frac{1}{H \times W} \sum_{i=1}^{H} \sum_{j=1}^{W} F_{i,j,c}, \quad c = 1, \dots, C.
\end{equation}

The descriptor then passed through a 1D convolution with kernel size \( k = 3  \), followed by a sigmoid activation function to generate channel-wise attention weights:

\begin{equation}
w = \sigma\big(\mathrm{Conv1D}_{k=3}(z)\big) \in \mathbb{R}^C.
\end{equation}

The final attention-refined feature map is then computed as:
\begin{equation}
F' = w \odot F
\end{equation}

where \( \odot \) denotes channel-wise multiplication. This procedure is independently applied to 
\( F_1, F_2, F_3, \) and \( F_4 \), yielding the recalibrated maps 
\( F'_1, F'_2, F'_3, \) and \( F'_4 \). By selectively amplifying informative channels while preserving feature dimensionality, ECA enhances the network's sensitivity to subtle anatomical variations, thereby strengthening multi-scale feature fusion in the decoder.

\subsubsection{Feature Bottleneck}
The feature bottleneck bridges the encoder and decoder by enriching the network’s deepest semantic features with multi-scale context before any upsampling. In convolutional networks, the spatial resolution usually decreases at each stage, making it more challenging to preserve large-scale features significantly. To address this, we incorporate the ASPP module at the final encoder output, aggregating multi-scale context without additional downsampling. Although our ASPP block is inspired by the architectural design of DeepLabV3+ \cite{chen2018deeplab}, it is explicitly built using clearly defined operations. Within the ASPP branches, we deliberately exclude batch normalization and separable convolutions, and also do not apply dropout inside ASPP to better preserve subtle details critical in grayscale, low-contrast ultrasound images. Let \( F'_4 \in \mathbb{R}^{16 \times 16 \times 576} \) be the ECA-refined output from the deepest encoder layer. The ASPP module processes this tensor through five parallel branches to capture multi-scale context. First, a point-wise \( (1 \times 1) \) convolution with ReLU activation preserves local details:

\begin{equation}
F_{1\times 1}
= \mathrm{ReLU}\!\left(
  \operatorname{Conv2D}_{1\times 1}\!\left( F'_4 \right)
\right)
\end{equation}

Simultaneously, three parallel \(3 \times 3\) convolutions with dilation rates \(d \in \{6,12,18\}\) are applied to \(F'_{4}\) to extract broader semantic context:

\begin{equation}
F_{d} = \mathrm{ReLU}\bigl(\operatorname{Conv2D}_{3\times 3,\,\mathrm{dilation}=d}(F'_{4})\bigr)\
\end{equation}

Additionally, a global pooling pathway compresses \(F'_4\)  through global average pooling (GAP), which is then projected using a \(1 \times 1\) convolution and ReLU before being bilinearly upsampled: 
\begin{equation}
F_{\mathrm{global}} =
\mathrm{Upsample}\!\left(
  \mathrm{ReLU}\!\left(
    \operatorname{Conv2D}_{1\times 1}\!\left(\mathrm{GAP}(F'_4)\right)
  \right)
\right)
\end{equation}

These five outputs are concatenated and fused through a final point-wise projection:

\begin{equation}
F_{\mathrm{ASPP}} =
\mathrm{ReLU}\!\left(
  \operatorname{Conv2D}_{1\times 1}\!\left(
    \mathrm{Concat}\!\left(F_{1\times 1},\,F_{6},\,F_{12},\,F_{18},\,F_{\mathrm{global}}\right)
  \right)
\right)
\end{equation}

This produces \( F_{\text{ASPP}} \in \mathbb{R}^{16 \times 16 \times 256} \), which preserves spatial resolution while integrating local and global contextual information. To introduce regularization while maintaining spatial consistency, we apply SpatialDropout with a dropout rate of 0.2 to the ASPP output. Unlike standard dropout, which randomly zeros individual activations, SpatialDropout removes entire feature maps during training, promoting channel-wise robustness \cite{tompson2015efficient}. We further refine the multi-scale representation using a custom CBAM attention block, i.e., a CBAM-inspired variant adapted for lightweight ultrasound segmentation. Combined with ASPP, this bottleneck stage enhances contextual feature aggregation prior to the BiomedCLIP-guided decoder fusion process. While inspired by the original CBAM architecture \cite{woo2018cbam}, the proposed variant employs a compact global spatial-context modulation strategy to reduce computational overhead while preserving contextual recalibration. In particular, the spatial attention pathway differs from the original CBAM formulation by operating on compressed global descriptors rather than dense spatial feature maps, making it more suitable for low-resolution bottleneck representations and real-time inference.
In the channel attention stage, we apply both global average pooling (GAP) and global max pooling (GMP) to capture spatial information. Each resulting vector is processed through a shared dense layer with ReLU activation. The outputs are then summed and passed through a second dense layer with sigmoid activation to produce channel-wise attention weights. These weights are subsequently broadcast and used to modulate the input feature map:

\begin{equation}
F_c \;=\; F_{\mathrm{ASPP}} \;\odot\; \sigma\!\left(
\mathrm{Dense}\!\left(
\mathrm{ReLU}\!\big(\mathrm{shared\_dense}(\mathrm{GAP}(F_{\mathrm{ASPP}}))\big)
+
\mathrm{ReLU}\!\big(\mathrm{shared\_dense}(\mathrm{GMP}(F_{\mathrm{ASPP}}))\big)
\right)\right).
\end{equation}

In the spatial attention stage, GAP and GMP are reapplied to channel-attended features \( F_c \), generating two pooled vectors. These vectors (each of length $C$) are concatenated to form a 2$C$-dimensional vector, reshaped into a \(1 \times 1 \times 2C\) tensor, which is then passed through a \(7 \times 7\) convolution with a single output channel and sigmoid activation to produce a scalar attention weight that uniformly modulates \(F_c\).

\begin{equation}
F_{\mathrm{CustomCBAM}}
= F_c \,\odot\,
\sigma\!\left(
  \operatorname{Conv2D}_{7\times 7}\!\left(
    \mathrm{Reshape}\!\left(
      \mathrm{Concat}\!\bigl(\mathrm{GAP}(F_c),\,\mathrm{GMP}(F_c)\bigr)
    \right)
  \right)
\right)
\end{equation}

\begin{figure*}[htbp]
  \centering
  \includegraphics[scale=0.22]{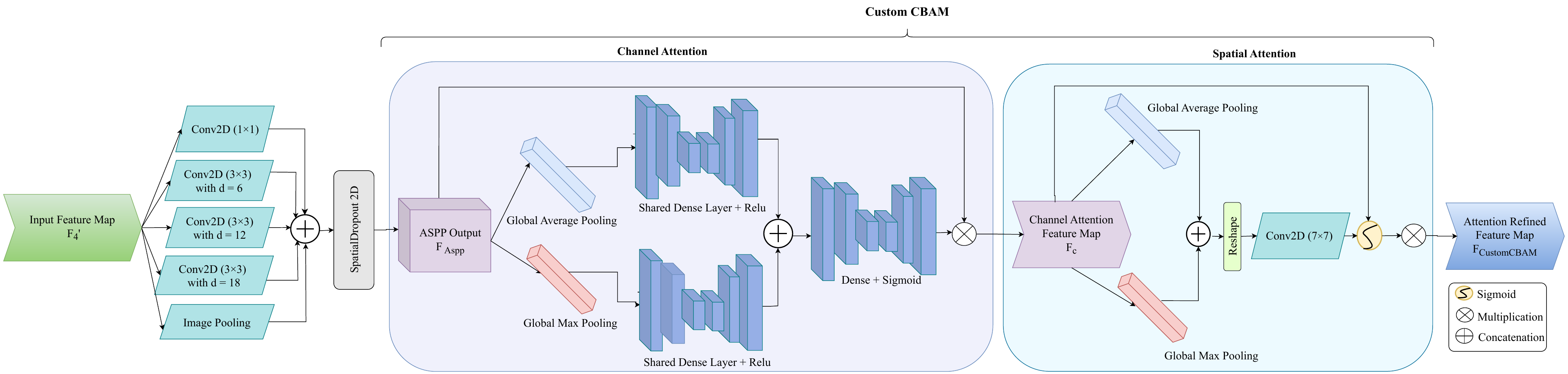}
  \captionsetup{justification=justified}

  \caption{Feature bottleneck module: the ECA-refined encoder output \(F'_4\) is processed by ASPP block with SpatialDropout to yield \(F_{\mathrm{ASPP}}\), which is then refined by custom CBAM to produce \(F_{\mathrm{CustomCBAM}}\) before decoding.}
  \label{fig:bottleneck_module}
\end{figure*}

This produces an output \(F_{\text{CustomCBAM}} \in \mathbb{R}^{16 \times 16 \times 256}\) that preserves spatial resolution while emphasizing informative regions across both spatial and channel dimensions. This enriched tensor is then forwarded to the decoder for precise mask reconstruction. Here, \hyperref[fig:bottleneck_module]{Figure~\ref*{fig:bottleneck_module}} depicts the bottleneck module, which enhances semantic representation before decoding.

\subsubsection{Decoder Pathway}
The decoder pathway reconstructs the final segmentation mask by progressively upsampling semantically enriched features from the bottleneck. Although the bottleneck employs an ASPP and a custom CBAM attention module to enhance semantic context, the decoder faces challenges in recovering fine-grained structural details, particularly in anatomically ambiguous or low-contrast ultrasound regions. 

To address this, we propose a three-stage hierarchical decoder architecture that integrates encoder skip connections, lightweight CBAM attention modules, and a BiomedCLIP-guided semantic refinement pathway. At each decoder stage, the feature map undergoes bilinear upsampling by a factor of two, followed by concatenation with the corresponding encoder features. These skip connections help restore spatial details that may have been lost during downsampling in the encoder.
After concatenation, the combined feature map passes through a lightweight CBAM block, which refines feature representation by applying both channel and spatial attention. This is followed by two convolutional layers, each equipped with ReLU activation, batch normalization, and dropout. These layers aim to sharpen structural boundaries and suppress irrelevant background features.

In the lightweight CBAM block, channel attention is computed by applying GAP to the input feature map \(F\) followed by two fully connected layers with ReLU and sigmoid activations to generate channel-wise weights, which modulate the input tensor. In compact form:
\begin{equation}
F_{\mathrm{ca}} = F \,\odot\, \sigma\!\bigl(\mathrm{Dense2}\bigl(\mathrm{ReLU}(\mathrm{Dense1}(\mathrm{GAP}(F)))\bigr)\bigr)
\end{equation}
Spatial attention is then calculated by applying GAP and GMP over channel-refined features, concatenated and reshaped to \(1 \times 1\) spatial dimensions, then passing through \(7\times7\) convolution (one output channel) with sigmoid to produce a scalar attention weight that uniformly modulates \(F_{\mathrm{ca}}\).

\begin{equation}
F_{\mathrm{LightweightCBAM}}
= F_{\mathrm{ca}} \,\odot\,
\sigma\!\left(
  \operatorname{Conv2D}_{7\times 7}\!\left(
    \mathrm{Reshape}\!\left(
      \mathrm{concat}\!\bigl(\mathrm{GAP}(F_{\mathrm{ca}}),\,\mathrm{GMP}(F_{\mathrm{ca}})\bigr)
    \right)
  \right)
\right)
\end{equation}

\begin{figure*}[!ht]
\centering
\includegraphics[scale=0.28]{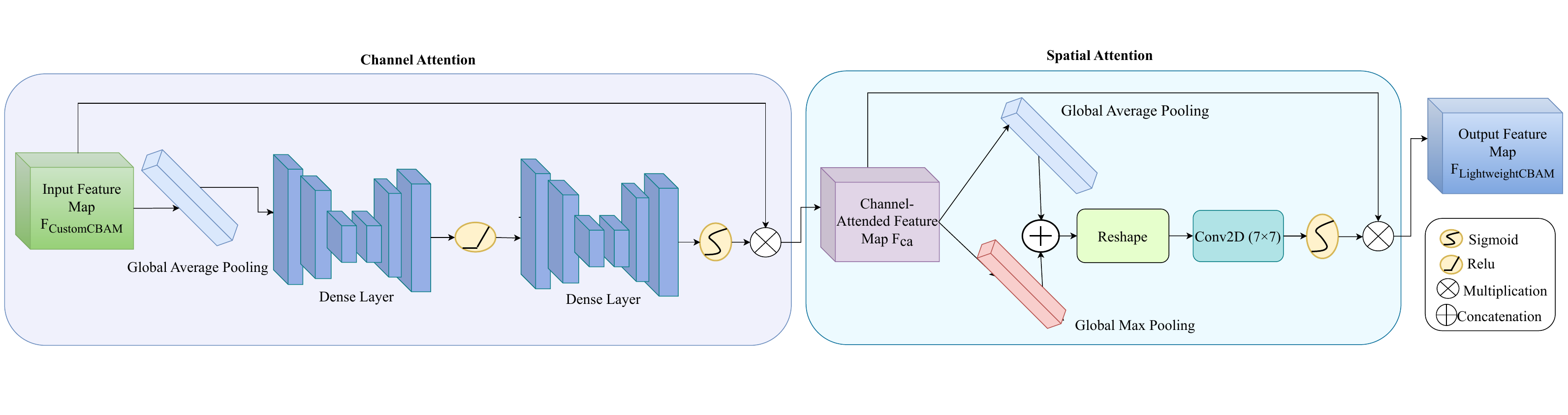}
\captionsetup{justification=justified}

\caption{Lightweight CBAM attention block: decoder feature maps first pass through channel attention (global average pooling $\rightarrow$ dense $\rightarrow$ relu $\rightarrow$ dense $\rightarrow$ sigmoid) and then spatial attention (pooled features concatenated $\rightarrow$ $7\times7$ convolution $\rightarrow$ sigmoid) to selectively refine and enhance informative regions before further decoding.}

\label{fig:cbam_block}
\end{figure*}

Together, these attention mechanisms refine spatial focus and enhance informative regions, facilitating better boundary delineation during decoding. \hyperref[fig:cbam_block]{Figure~\ref*{fig:cbam_block}} illustrates the architecture of the lightweight CBAM attention block used to enhance channel and spatial feature representation within the decoder.

To further address semantic ambiguity, particularly at lower decoder resolutions, we incorporate a BiomedCLIP-guided Gated Fusion (BGF) module that leverages guidance exclusively from BiomedCLIP’s vision encoder without text prompts, thereby keeping the framework fully image-based and lightweight. This component introduces global biomedical semantic priors into the decoder using projected BiomedCLIP image embeddings. Let, \( f_{\text{BiomedCLIP}} \in \mathbb{R}^C \) denote the BiomedCLIP image embedding vector, representing global biomedical semantics derived exclusively from the vision encoder and \( F_d \in \mathbb{R}^{H \times W \times C} \) represent the decoder feature map.  
To incorporate global context into the spatial decoder features, the BiomedCLIP vector is first projected into the decoder feature space using a fully connected layer with ReLU activation. The resulting output is reshaped to a spatial size of 
\(1 \times 1\) and bilinearly upsampled to match the spatial resolution of the decoder features:

\begin{equation}
f' = \text{Upsample} \big( \text{Reshape} ( \text{ReLU}(\text{Dense}(f_{\text{BiomedCLIP}}))) \big) \in \mathbb{R}^{H \times W \times C}
\end{equation}
The upsampled feature \( f' \) is fused with the decoder feature map \( F_d \) using a fixed weighted sum, where global semantic features are given higher priority (80\%) over local decoder features (20\%). 

\begin{equation}
F_{\text{fused}} = \alpha \, f' + (1 - \alpha)\, F_d
\end{equation}

where $\alpha \in [0,1]$ denotes the fixed fusion coefficient that controls the relative contribution of the global semantic embedding and the local decoder features. In the proposed configuration, $\alpha = 0.8$. The fixed weighting (0.8 for global semantic embedding and 0.2 for local decoder features) is adopted to maintain stable semantic guidance while preserving the lightweight design of the framework. In thyroid ultrasound imaging, where contrast is low and anatomical boundaries are often ambiguous, stronger global semantic guidance helps regularize decoder activations while local refinement is preserved through skip-connected decoder features. To validate this design choice, we performed repeated-run sensitivity analysis across multiple fixed $\alpha$ values and a learnable adaptive gating variant (Supplementary Tables S1–S2). Results are reported as mean ± standard deviation over five independent runs, together with one-sided Wilcoxon signed-rank significance testing. The results consistently show that $\alpha = 0.8$ provides the most favorable balance within the tested settings between overlap quality (Dice/IoU) and boundary fidelity (HD95) across both TN3K and DDTI. Compared with adaptive gating, the proposed fixed-ratio strategy achieves comparable or stronger performance while avoiding additional adaptive-gating parameterization (8.63M vs. 8.55M parameters). These findings support the proposed fixed fusion design as a stable, lightweight, and reproducible configuration.

\begin{figure*}[!h]
  \centering
  \includegraphics[scale=0.32]{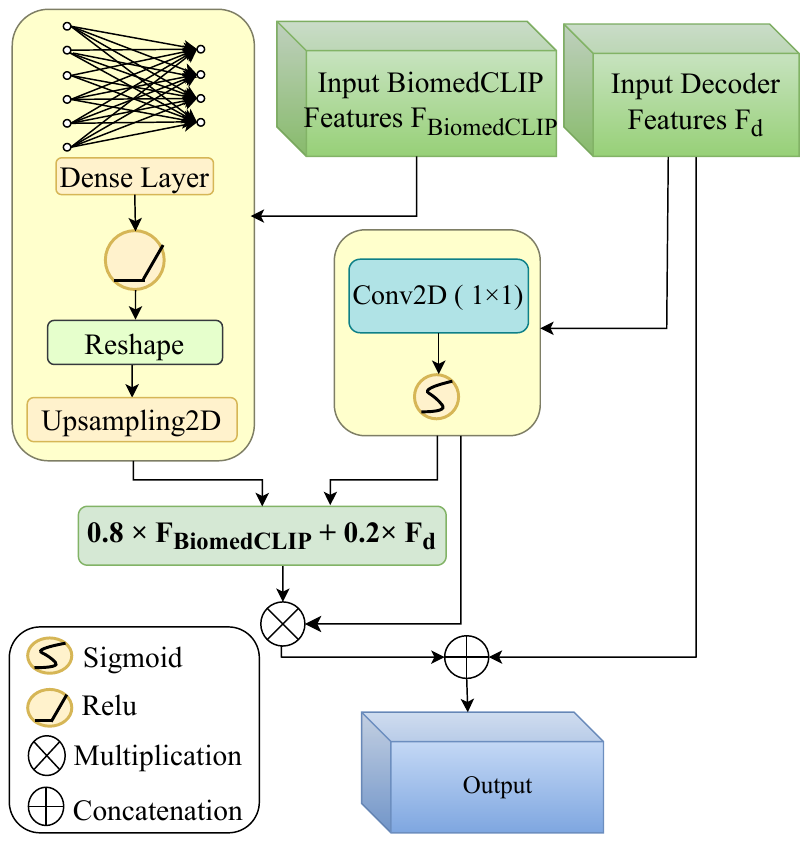}
  \captionsetup{justification=justified}

  \caption{BiomedCLIP-guided Gated Fusion (BGF) module: global biomedical image embeddings are projected, reshaped, and bilinearly upsampled, then combined with decoder features using weighted addition and refined with spatial attention gating before concatenation.}
  \label{fig:biomedclip_fusion}
\end{figure*}

A spatial attention map \( A \in \mathbb{R}^{H \times W \times C} \) is computed from decoder features using a \( 1 \times 1 \) convolution followed by sigmoid activation:

\begin{equation}
A = \sigma \big( \text{Conv2D}_{1 \times 1}(F_d) \big)
\end{equation}
This attention map is used to recalibrate the fused features using element-wise multiplication:
\begin{equation}
F_{\mathrm{gated}} = A \odot F_{\mathrm{fused}}
\end{equation}
To preserve both global semantic guidance and local spatial information, the gated features are concatenated with the original decoder output:
\begin{equation}
F_{\text{out}} = \text{Concat}(F_d, F_{\text{gated}}) \in \mathbb{R}^{H \times W \times 2C}
\end{equation}
This BiomedCLIP fusion is selectively applied at decoder resolutions of \(32 \times 32\) and \(64 \times 64\), where semantic ambiguity is most significant and global guidance is crucial. It is intentionally excluded at the  \(128 \times 128\) resolution to preserve high-frequency spatial details and prevent over-smoothing caused by excessive global semantic influence. \hyperref[fig:biomedclip_fusion]{Figure~\ref*{fig:biomedclip_fusion}} illustrates the BiomedCLIP-guided semantic fusion process, where projected global image embeddings are selectively integrated with decoder representations through weighted fusion and spatial attention refinement.

Finally, the decoder output is passed through a \(1 \times 1\) convolutional layer with sigmoid activation, followed by a bilinear upsampling to the original image resolution, yielding the final segmentation mask.


\subsection{Experimental Parameters}
To train the proposed architecture, each dataset was first divided into training, validation, and test subsets according to the dataset-specific protocol, and data augmentation was applied only to the training subset. For the TG3K, DDTI, and PKTN datasets, an 80:10:10 split was applied, allocating 80\% of the images for training, 10\% for validation, and 10\% for testing. For the TN3K dataset, the original test split remained unchanged, and only the training set was further divided into training and validation subsets using an 80:20 ratio, allocating 80\% of the images for training, 20\% for validation. All images across the datasets were uniformly resized to \(256 \times 256\) pixels and normalized to the \([0, 1]\) range to ensure consistent input dimensions. A batch size of 32 was used throughout the training process to balance computational efficiency and hardware limitations. The model employed MobileNetV2 as the encoder backbone and was initialized with ImageNet-pretrained weights to leverage transfer learning. Training was conducted for 150 epochs using the Adam optimizer with an initial learning rate of \(1 \times 10^{-4}\). A cosine annealing scheduler was utilized to gradually reduce the learning rate over time, complemented by a ReduceLROnPlateau callback with a factor of 0.1 and a patience of 10 epochs, which further decreased the learning rate when the validation loss plateaued. Early stopping was also implemented, with a patience of 20 epochs, to prevent overfitting. All experiments were carried out on an NVIDIA Tesla P100 PCIe GPU with 16 GB of VRAM, using Tensorflow as the deep learning framework. The model has 8,550,720 (32.62 MB) trainable parameters, which remain consistent across all datasets due to the fixed architecture. To address region-overlap errors, foreground--background imbalance, and difficult pixels, we employed a hybrid objective combining DICE loss, IoU loss, and Focal Tversky loss. DICE loss emphasizes region overlap, IoU loss reinforces global intersection quality, and Focal Tversky loss provides additional emphasis on class imbalance and difficult foreground regions. The hybrid loss is defined as:

\begin{equation}
\mathcal{L}_{\mathrm{hybrid}}
=
\lambda_{1}\mathcal{L}_{\mathrm{DICE}}
+
\lambda_{2}\mathcal{L}_{\mathrm{IoU}}
+
\lambda_{3}\mathcal{L}_{\mathrm{FocalTversky}},
\end{equation}

where $\lambda_{1}$, $\lambda_{2}$, and $\lambda_{3}$ denote the weighting coefficients of the DICE, IoU, and Focal Tversky losses, respectively. In this study, the coefficients were fixed as $\lambda_{1}=0.4$, $\lambda_{2}=0.2$, and $\lambda_{3}=0.4$ and were maintained consistently across all experiments.

The DICE loss is given by:

\begin{equation}
\mathcal{L}_{\mathrm{DICE}}
=
1-
\frac{
2\sum_{i=1}^{T}p_i g_i+\epsilon
}{
\sum_{i=1}^{T}p_i+\sum_{i=1}^{T}g_i+\epsilon
}.
\end{equation}

The IoU loss is calculated as follows:

\begin{equation}
\mathcal{L}_{\mathrm{IoU}}
=
1-
\frac{
\sum_{i=1}^{T}p_i g_i+\epsilon
}{
\sum_{i=1}^{T}p_i+
\sum_{i=1}^{T}g_i-
\sum_{i=1}^{T}p_i g_i+\epsilon
}.
\end{equation}

where $p_i$ and $g_i$ represent the predicted and ground-truth values for pixel $i$, $T$ denotes the total number of pixels, and $\epsilon$ is a smoothing constant for numerical stability. In the implementation, the smoothing constant was set to $10^{-6}$ for the DICE and IoU losses and to $1$ for the Tversky index. %

The Focal Tversky loss is defined using the Tversky index $TI_c$ for class $c$:

\begin{equation}
\mathcal{L}_{FocalTversky} = \sum_c (1 - TI_c)^\gamma, \quad
TI_c = \frac{\sum_{i=1}^T p_{ic} g_{ic} + \epsilon}{\sum_{i=1}^T p_{ic} g_{ic} + \alpha \sum_{i=1}^T p_{ic} (1 - g_{ic}) + \beta \sum_{i=1}^T (1 - p_{ic}) g_{ic} + \epsilon}
\end{equation}

Here, $p_{ic}$ and $g_{ic}$ denote the predicted and ground-truth probabilities for pixel $i$ and class $c$, respectively. The coefficients $\alpha$ and $\beta$ control the penalties for false positives and false negatives, respectively. The false-positive and false-negative coefficients were set to $\alpha=0.4$ and $\beta=0.6$, respectively, with the focusing parameter fixed at $\gamma=0.85$ across all experiments.

\subsection{Evaluation Metrics}
To evaluate the performance of our segmentation model, we employed six evaluation metrics that capture region-wise accuracy, structural overlap, and boundary precision. Accuracy measures the overall proportion of correctly classified pixels. Precision quantifies the correctness of positive predictions, while recall evaluates the model’s ability to detect all actual positive pixels. IoU and DICE both assess spatial overlap between predicted and ground truth masks, with DICE being more sensitive to small regions. The 95th percentile Hausdorff Distance (HD95) evaluates boundary-related spatial discrepancy between the predicted and ground-truth segmentation masks while reducing sensitivity to extreme outliers. HD95 was computed for each test image in the resized 256 × 256-pixel space using the 95th-percentile Euclidean distance between the foreground pixels of the predicted and ground-truth segmentation masks, with predictions thresholded at 0.5. The reported HD95 values were averaged across the test set and expressed in pixels (px). The pixel-based metrics are formally defined as follows:

\begin{equation}
\text{Accuracy} = \frac{TP + TN}{TP + TN + FP + FN} \times 100
\end{equation}

\begin{equation}
\text{Precision} = \frac{TP}{TP + FP} \times 100
\end{equation}

\begin{equation}
\text{Recall} = \frac{TP}{TP + FN} \times 100
\end{equation}

\begin{equation}
\text{IoU} = \frac{TP}{TP + FP + FN} \times 100
\end{equation}

\begin{equation}
\text{DICE} = \frac{2 \times TP}{2 \times TP + FP + FN} \times 100
\end{equation}
Where true positives (\(TP\)) represent the number of pixels correctly identified by the model as belonging to the target structure (e.g., thyroid gland or nodule), and true negatives (\(TN\)) are the background pixels correctly identified as non-target by the model. False positives (\(FP\)) are background pixels that the model incorrectly labels as part of the target structure, while false negatives (\(FN\)) are pixels that belong to the ground truth target region but are missed by the model. These definitions are computed on a per-pixel basis across the segmentation masks and directly reflect the ability of our proposed ThyCLIPNet model to accurately delineate anatomical structures in complex ultrasound images. 

\section{Experimental Analysis}
\label{sec:results}
This section presents a comprehensive evaluation of the proposed ThyCLIPNet model through both quantitative and qualitative analyses. We compare performance against state-of-the-art (SOTA) methods across multiple thyroid ultrasound datasets using standardized metrics. Additionally, we conduct ablation studies to assess the contribution of each architectural component and analyze trade-offs in computational efficiency. Finally, we employ Grad-CAM visualizations to interpret the model's attention behavior.

\subsection{Quantitative Comparison with SOTA Methods}
We evaluated the performance of ThyCLIPNet across four publicly available thyroid ultrasound image segmentation datasets: TG3K, TN3K, DDTI, and PKTN. Competing methods included: Coarse for Fine \cite{10856358}, Weakly-supervised Framework \cite{ALY2025109669}, Adapting Vision \cite{zhang2025adaptingvisionfoundationmodels}, DW‐Swin \cite{10847842}, TRFE+ \cite{GONG2023106389}, BFG \& MSF‐Net \cite{liu2024bfg}, CC‐SAMl \cite{Gowda2024CCSAMSW}, STU$^3$Net \cite{deng2024stu3net}, SHAN \cite{Zha_SHAN_MICCAI2024}, BPAT‐UNet \cite{BI2023107614}, GDSSA-Net \cite{farooq2025gdssa}, MCFU-Net \cite{liu2025prediction}, SAMTNU \cite{chen2025samtnu}, UC-Seg \cite{huang2025uncertainty}, HFA-UNet \cite{li2025hfa}, CS-Net \cite{wang2026medical}, PEW-SegDiff \cite{bi2025pew}, LMSFE-SAM \cite{jing2025medical}, BTNet \cite{li2023novel}, CRSANet \cite{sun2024crsanet}, FCG-Net \cite{shao2023fcg}, DBUNet \cite{zhu2025dual}, FMF-Net \cite{liu2024thyroid}, DCDT \cite{chi2025dynamic}, GCAESeg \cite{kang2025gcaeseg} and other representative state-of-the-art approaches. \hyperref[tab:comparison_bold]{Table~\ref*{tab:comparison_bold}} presents a detailed breakdown of performance across six key metrics, namely DICE, IoU, accuracy, recall, precision, and HD95. Since official code, pretrained checkpoints, or fully matched evaluation protocols were not consistently available for all competing methods, the comparative values reported here are primarily drawn from the corresponding original publications. Accordingly, these results are interpreted as reference benchmarks against published methods.

\begin{table*}[ht]
\captionsetup{labelformat=empty}
\caption{\textbf{Table~\ref*{tab:comparison_bold}.} Comparison of segmentation effects between the proposed and SOTA methods across all datasets.}
\label{tab:comparison_bold}
\resizebox{\textwidth}{!}{%
  \begin{tabular}{l l c c c c c c}
    \hline
    \textbf{Dataset} & \textbf{Methods} & \textbf{DICE (\%) $\uparrow$} & \textbf{IoU (\%) $\uparrow$} & \textbf{Accuracy (\%) $\uparrow$} & \textbf{Recall (\%) $\uparrow$} & \textbf{Precision (\%) $\uparrow$} & \textbf{HD95 $\downarrow$} \\
    \hline
    \multirow{9}{*}{TG3K \cite{wunderling2017comparison}}
     
     & Coarse for Fine \cite{10856358}
       & 80.67$\pm$1.09 & \textemdash & \textemdash & \textemdash & \textemdash & 16.63$\pm$0.84 \\
       & Weakly‐supervised Framework \cite{ALY2025109669}
       & 85.92$\pm$1.53
       & \textemdash
       & 93.00
       & 86.00
       & \cellcolor{bestcolor}{91.00}
       & 13.09$\pm$0.43 \\
     & Adapting Vision \cite{zhang2025adaptingvisionfoundationmodels}
       & 83.04 & \textemdash & \textemdash & \textemdash & \textemdash & 12.55 \\
     & SAMTNU \cite{chen2025samtnu}
       & 83.87
       & \cellcolor{bestcolor}{73.79}

       & \cellcolor{bestcolor}{97.64}
       & 85.17
       & \textemdash
       & 23.98 \\
     & DCDT\cite{chi2025dynamic}
       & 82.04\,$\pm$\,1.11
       & 69.55\,$\pm$\,1.85
       & \textemdash
       & \textemdash
       & \textemdash
       & 26.68\,$\pm$\,0.74 \\
     
     & FMF-Net \cite{liu2024thyroid}
       & \cellcolor{bestcolor}92.20

       & \textemdash
       & \textemdash
       & \cellcolor{bestcolor}88.90
       & \textemdash
       & \cellcolor{bestcolor}\textbf{0.94} \\
     
    \rowcolor{proposedcolor} & ThyCLIPNet-Proposed
       & \textbf{96.22\,$\pm$\,0.10} & \textbf{92.72\,$\pm$\,0.10} & \textbf{99.33\,$\pm$\,0.00} & \textbf{96.80\,$\pm$\,0.20} & \textbf{95.64\,$\pm$\,0.20} & 3.75\,$\pm$\,0.16 \\
    \hline
    \multirow{25}{*}{TN3K \cite{gong2023thyroid}}
     
     & Coarse for Fine \cite{10856358}
       & 79.07$\pm$1.65 & \textemdash & \textemdash & \textemdash & \textemdash & 17.33$\pm$1.17 \\
       & Weakly‐supervised Framework \cite{ALY2025109669}
       & 84.81\,$\pm$\,1.47
       & \textemdash
       & 91.00
       & 86.00
       & 89.00
       & 14.16\,$\pm$\,1.25 \\
       & Striving for Simplicity \cite{chen2024striving}
       & 84.17
       & 75.70
       & \textemdash
       & \textemdash
       & \textemdash
       & \textemdash \\
       & CC‐SAMI \cite{Gowda2024CCSAMSW}
       & 85.20$\pm$16.93 & 77.89$\pm$19.75 & 97.22$\pm$4.58 & \cellcolor{bestcolor}88.56$\pm$19.88 & \textemdash & \textemdash \\
     & Adapting Vision \cite{zhang2025adaptingvisionfoundationmodels}
       & 86.01 & \textemdash & \textemdash & \textemdash & \textemdash & 15.43 \\
       
     & DW‐Swin \cite{10847842}
       & 82.26 & 73.00 & \textemdash & \textemdash & \textemdash & \textemdash \\
     & TRFE+ \cite{GONG2023106389}
       & 83.30$\pm$0.26 & 71.38$\pm$0.43 & 97.04$\pm$0.10 & \textemdash & \textemdash & 13.23$\pm$0.63 \\
     & BFG\&MSF‐Net \cite{10543017}
       & \cellcolor{bestcolor}86.82 & \cellcolor{bestcolor}\textbf{78.73} & \textemdash & 87.48 & 88.72 & \textemdash \\
     
     & STU$^3$Net \cite{deng2024stu3net}
       & 83.68 & 74.16 & 97.55 & \textemdash & \textemdash & 9.30 \\
     & SHAN \cite{Zha_SHAN_MICCAI2024}
       & 84.61$\pm$0.04 & 73.59$\pm$0.16 & 96.73$\pm$0.06 & \textemdash & \textemdash & \cellcolor{bestcolor}\textbf{4.05$\pm$0.27} \\
     & BPAT‐UNet \cite{BI2023107614}
       & 83.64 & 71.87 & 97.22 & 85.57 & 82.94 & \textemdash \\

     & DCDT \cite{chi2025dynamic}
       & 81.21\,$\pm$\,1.32
       & 68.35\,$\pm$\,2.18
       & \textemdash
       & \textemdash
       & \textemdash
       & 23.57\,$\pm$\,1.91 \\
       & GDSSA‐Net \cite{farooq2025gdssa}
       & 84.27
       & 73.03
       & 96.25
       & 81.63
       & 87.76
       & \textemdash \\
     & UC-Seg \cite{huang2025uncertainty}
       & 82.17
       & 73.34
       & \textemdash
       & \textemdash
       & \textemdash
       & 4.33 \\

       & LMSFE-SAM\cite{jing2025medical}
       & 83.15
       & 72.27
       & \cellcolor{bestcolor}\textbf{98.67}
       & 79.72
       & \textemdash
       & 17.35 \\

       & MCFU-net \cite{liu2025prediction}
       & 85.22\,$\pm$\,0.34
       & 74.25\,$\pm$\,0.53
       & 96.85\,$\pm$\,0.08
       & 79.86\,$\pm$\,0.97
       & \cellcolor{bestcolor}\textbf{92.14\,$\pm$\,0.66}

       & 23.45\,$\pm$\,1.65 \\
     & HFA-UNet\cite{li2025hfa}
       & 84.58
       & \textemdash
       & \textemdash
       & \textemdash
       & \textemdash
       & 38.85 \\
     & PEW-SegDiff\cite{bi2025pew}
       & 78.64
       & 74.29
       & 91.94
       & 78.81
       & \textemdash
       & 14.61 \\
     & CS-Net\cite{wang2026medical}
       & 86.47
       & \textemdash
       & \textemdash
       & 86.30
       & 86.51
       & \textemdash \\
     & SAMTNU \cite{chen2025samtnu}
       & 84.65
       & 75.14
       & 97.32
       & 88.37
       & \textemdash
       & 27.82 \\
     & GCAESeg\cite{kang2025gcaeseg}
       & \textemdash
       & 69.39
       & \textemdash
       & \textemdash
       & \textemdash
       & \textemdash \\
    \rowcolor{proposedcolor} & ThyCLIPNet-Proposed
       & \textbf{87.58\,$\pm$\,0.30} & 77.91\,$\pm$\,0.50 & 96.94\,$\pm$\,0.10 & \textbf{89.47\,$\pm$\,0.40} & 85.78\,$\pm$\,0.50 & 16.38\,$\pm$\,0.92 \\
    \hline
    \multirow{22}{*}{DDTI \cite{pedraza2015open}}
     & Adapting Vision \cite{zhang2025adaptingvisionfoundationmodels}
       & \cellcolor{bestcolor}81.52 & \textemdash & \textemdash & \textemdash & \textemdash & 26.68 \\
       & DBUNet\cite{zhu2025dual}
       & 79.90
       & \textemdash
       & 95.76
       & 81.44
       & 78.59
       & \textemdash \\
       & TRFE+ \cite{GONG2023106389}
       & 75.37$\pm$2.14 & 60.47$\pm$1.08 & 93.18$\pm$0.26 & \textemdash & \textemdash & 24.60$\pm$1.23 \\
       
       & BFG\&MSF‐Net \cite{10543017}
       & 81.02 & \cellcolor{bestcolor}69.35 & \textemdash & 81.19 & \cellcolor{bestcolor}\textbf{83.64} & \textemdash \\
     
       & FCG‐Net\cite{shao2023fcg}
       & 80.42
       & \textemdash
       & 95.40
       & 82.70
       & 82.36
       & \textemdash \\
      & DCDT \cite{chi2025dynamic}
       & 79.01\,$\pm$\,2.33
       & 65.31\,$\pm$\,2.07
       & \textemdash
       & \textemdash
       & \textemdash
       & 28.17\,$\pm$\,2.31 \\ 
     & CRSANet\cite{sun2024crsanet}
       & 74.26
       & \textemdash
       & \cellcolor{bestcolor}\textbf{97.11}
       & 75.31
       & 75.92
       & \textemdash \\

     & MCFU-net \cite{liu2025prediction}
       & 78.21\,$\pm$\,0.77
       & 64.23\,$\pm$\,1.05
       & 93.76\,$\pm$\,0.29
       & 77.89\,$\pm$\,1.75
       & 78.58\,$\pm$\,1.15
       & 24.78\,$\pm$\,1.55 \\
     & BTNet\cite{li2023novel}
       & 75.70
       & 65.40
       & \textemdash
       & \textemdash
       & \textemdash
       & \textemdash \\

     & GDSSA‐Net \cite{farooq2025gdssa}
       & 79.85
       & 68.96
       & 94.81
       & \cellcolor{bestcolor}83.03

       & 81.10
       & \textemdash \\

     & UC-Seg\cite{huang2025uncertainty}
       & 62.00
       & 50.96
       & \textemdash
       & \textemdash
       & \textemdash
       & \cellcolor{bestcolor}\textbf{5.54} \\
     & DW‐Swin \cite{10847842}
       & 78.64 & 67.93 & \textemdash & \textemdash & \textemdash & \textemdash \\
    \rowcolor{proposedcolor} & ThyCLIPNet-Proposed
       & \textbf{84.73\,$\pm$\,0.50} & \textbf{73.51\,$\pm$\,0.70} & 95.26\,$\pm$\,0.20 & \textbf{89.25\,$\pm$\,1.70} & 80.69\,$\pm$\,1.80 & 18.23\,$\pm$\,1.10 \\
    \hline
    \multirow{1}{*}{PKTN \cite{sun2025clip}}
     & GCAESeg\cite{kang2025gcaeseg}
       & \textemdash
       & \cellcolor{bestcolor}\textbf{78.08}
       & \textemdash
       & \textemdash
       & \textemdash
       & \textemdash \\
    \rowcolor{proposedcolor} & ThyCLIPNet-Proposed
       & \textbf{80.70\,$\pm$\,0.40}
       & 67.64\,$\pm$\,0.50
       & \textbf{98.67\,$\pm$\,0.00}
       & \textbf{80.22\,$\pm$\,1.80}
       & \textbf{81.25\,$\pm$\,2.00}
       & \textbf{10.86\,$\pm$\,0.46} \\
    \hline
  \end{tabular}%
}
\vspace{0.2cm}
\parbox{\textwidth}{%
  \footnotesize
  \textbf{Color Legend:} 
  \colorbox{proposedcolor}{\phantom{XX}} Proposed method, 
  \colorbox{bestcolor}{\phantom{XX}} Best non-proposed method for each metric; 
  \textbf{Bold}: Overall best performance including proposed method; 
  ($\uparrow$: higher is better, $\downarrow$: lower is better). Missing entries (—) indicate
values not reported in the original papers.
}
\end{table*}

Across all four datasets, ThyCLIPNet consistently delivers competitive performance, achieving the highest or near-highest scores in DICE, IoU, and recall, while maintaining strong accuracy and lower boundary error (HD95). Examining dataset-specific trends, the model performs particularly well on the TG3K dataset, where it attains the best overall overlap performance (highest DICE and IoU) and the highest reported accuracy, recall, and precision among methods that report these metrics. In particular, compared with Adapting Vision, ThyCLIPNet improves DICE by over 13 percentage points (pp) and reports a lower HD95. Although FMF-Net \cite{liu2024thyroid} reports the lowest HD95 on TG3K, ThyCLIPNet attains substantially higher overlap and sensitivity, demonstrating a more balanced trade-off between region coverage and clinically meaningful completeness, while maintaining competitive boundary accuracy.

Similarly, on the TN3K dataset, ThyCLIPNet achieves the highest DICE and recall, indicating superior region overlap and sensitivity compared with other advanced architectures. Although BFG \& MSF-Net \cite{10543017} achieves higher IoU and precision, ThyCLIPNet attains higher Dice and recall, indicating more complete nodule coverage with fewer missed regions. Methods optimized for boundary sharpness, such as SHAN \cite{Zha_SHAN_MICCAI2024} and UC-Seg \cite{huang2025uncertainty}, report lower HD95 values, indicating more precise edge localization. However, their lower DICE and IoU scores reveal weaker region-level consistency, suggesting that although boundaries are well defined, overall gland coverage and overlap accuracy remain suboptimal. Conversely, approaches with larger HD95 values (e.g., SAMTNU \cite{chen2025samtnu}, 27.82, and MCFU-Net \cite{liu2025prediction}, 23.45) exhibit reduced boundary robustness. Overall, ThyCLIPNet provides the best balance of overlap and sensitivity on TN3K, while maintaining competitive precision and accuracy, resulting in a more consistent and well-rounded performance profile.

For the smaller and noisier DDTI dataset, achieving the highest DICE, superior overlap and sensitivity highlights ThyCLIPNet’s strong performance, demonstrating consistent region coverage and reduced under-segmentation under challenging imaging conditions. Although BFG \& MSF-Net \cite{10543017} reports higher precision, its lower DICE and IoU suggest a more conservative prediction strategy that prioritizes fewer false positives at the expense of overall region overlap. UC-Seg \cite{huang2025uncertainty} achieves the lowest HD95, yet its substantially lower DICE and IoU indicate severe under-segmentation and poor structural completeness. In contrast, several methods, such as TRFE+ \cite{GONG2023106389}, MCFU-Net \cite{liu2025prediction}, and DCDT \cite{chi2025dynamic}, exhibit significantly larger boundary errors, whereas ThyCLIPNet maintains a reduced HD95 while preserving region-level accuracy.

\begin{figure*}[htbp]
    \centering

    \begin{subfigure}[b]{\textwidth}
        \centering
        \includegraphics[width=1.0\linewidth]{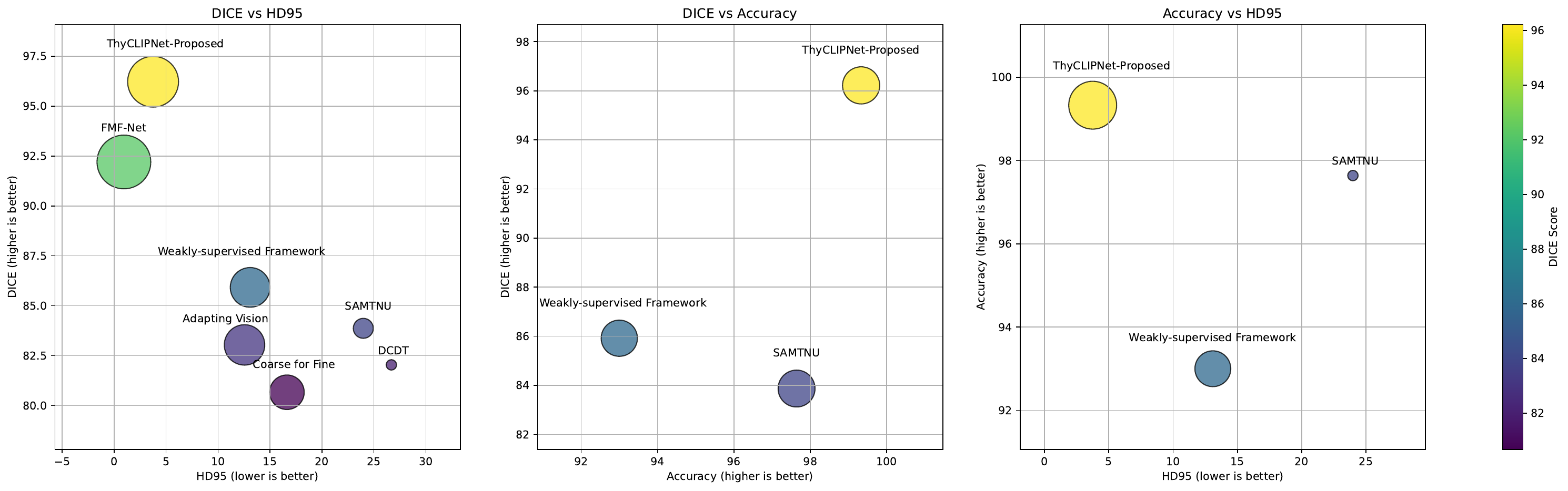}
    \end{subfigure}
    \vspace{1em}

    \begin{subfigure}[b]{\textwidth}
        \centering
        \includegraphics[width=1.0\linewidth]{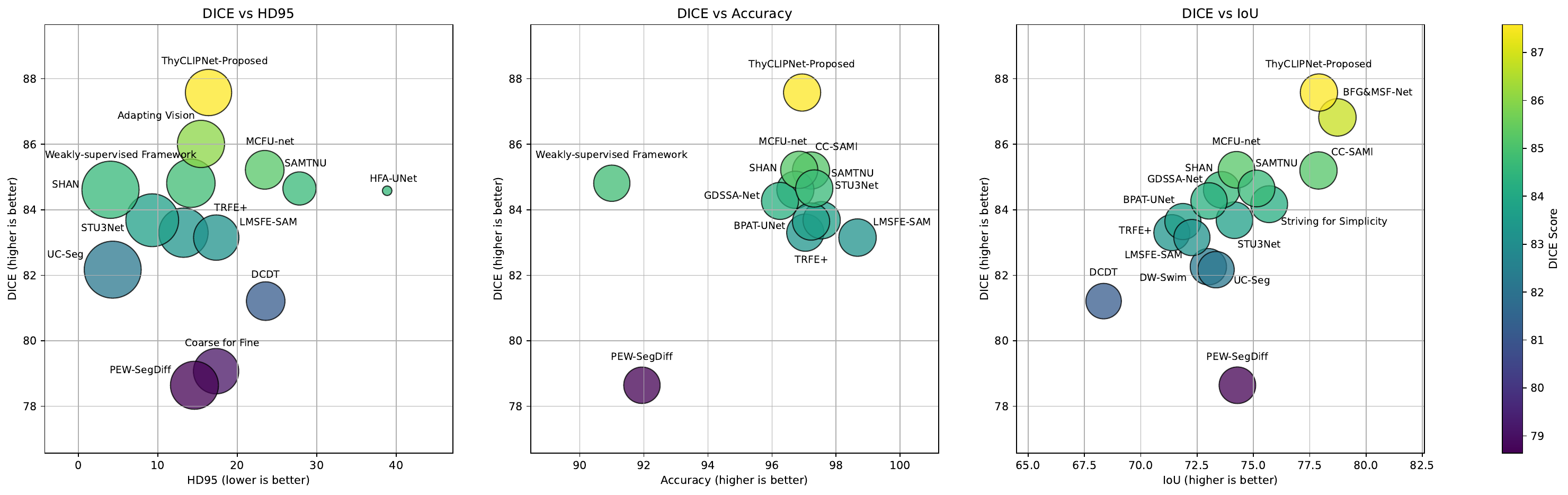}
    \end{subfigure}
    \vspace{1em}

    \begin{subfigure}[b]{\textwidth}
        \centering
        \includegraphics[width=1.0\linewidth]{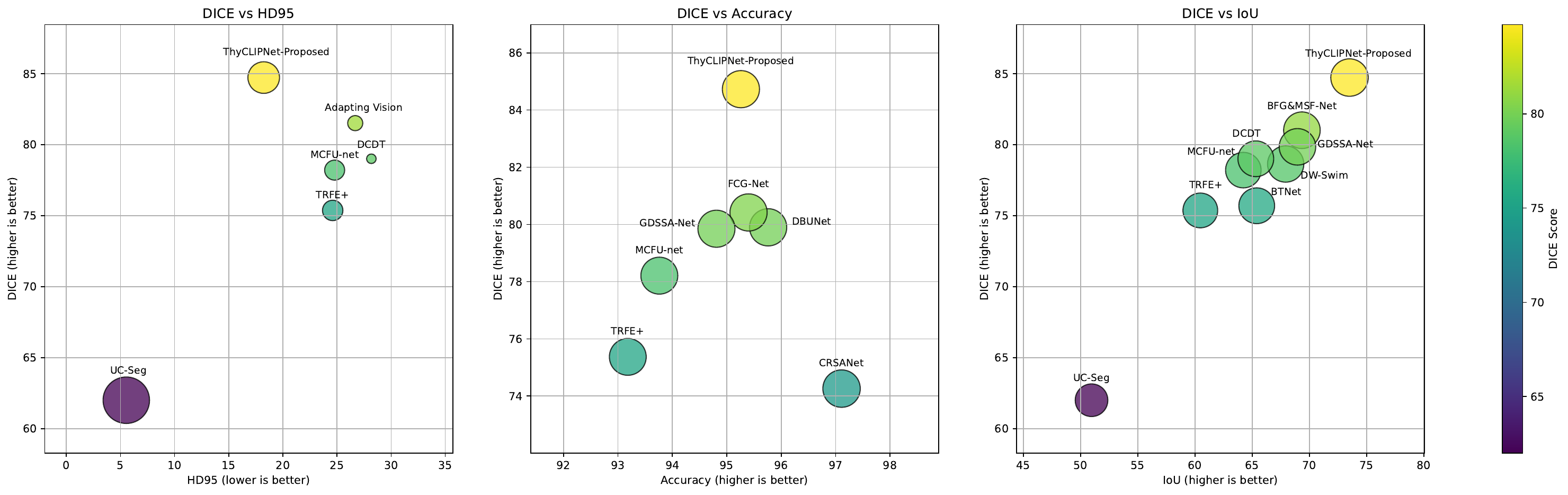}
    \end{subfigure}
\captionsetup{justification=justified}

    \caption{Multi‐metric bubble‐plot comparison of ThyCLIPNet versus state‐of‐the‐art methods on three thyroid ultrasound datasets (TG3K: top row; TN3K: middle row; DDTI: bottom row). The left column shows DICE versus HD95 (bubble size $\propto 1/\mathrm{HD95}$); the center column shows DICE versus Accuracy (bubble size $\propto \mathrm{Accuracy}$); and the right column shows dataset‐specific comparisons: Accuracy versus HD95 for TG3K (bubble size $\propto 1/\mathrm{HD95}$), DICE versus IoU for TN3K and DDTI (bubble size $\propto \mathrm{IoU}$). Bubble color represents the DICE score.
}
    \label{fig:bubble_comparison}
\end{figure*}

Finally, on the PKTN dataset, where comprehensive SOTA benchmarks are limited, ThyCLIPNet establishes strong reference baselines across all major evaluation metrics. While prior work, such as GCAESeg \cite{kang2025gcaeseg}, reports only partial metrics, the incomplete reporting limits direct comparison across all six metrics. On the contrary, ThyCLIPNet provides a complete metric profile, demonstrating its ability to produce accurate, stable, and boundary-consistent segmentations. These results highlight ThyCLIPNet’s robustness and strong generalization to real-world, clinically diverse thyroid ultrasound data, positioning it as a reliable benchmark for future research on this dataset. The bubble plot in \hyperref[fig:bubble_comparison]{Figure~\ref*{fig:bubble_comparison}} visually reinforces these findings, illustrating that ThyCLIPNet attains the highest DICE and recall scores, while delivering competitive accuracy and precision with low boundary error (HD95).

\subsection{Qualitative Comparison and Error Analysis}
\begin{figure*}[htbp]    \centerline{\includegraphics[width=\textwidth]{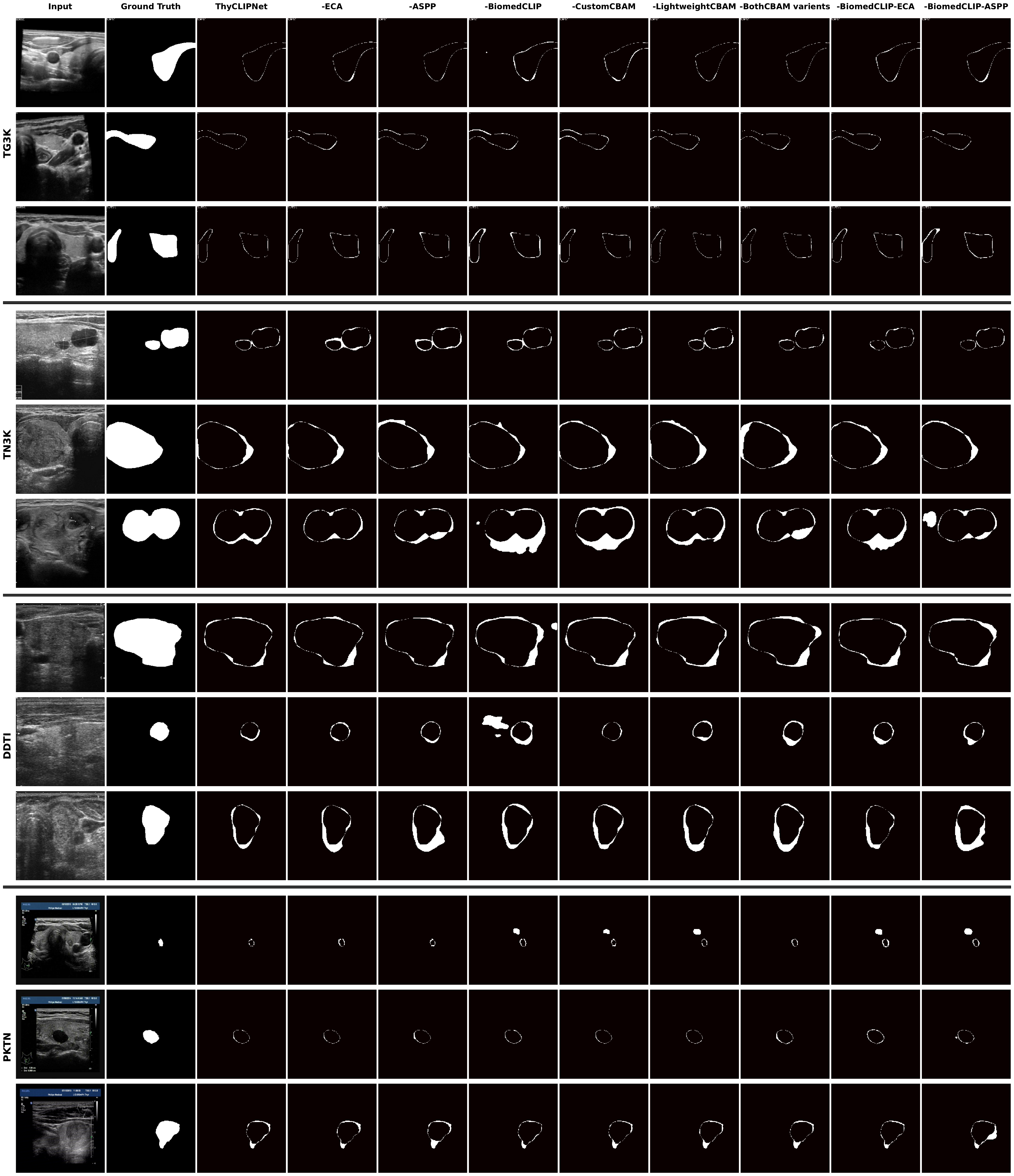}}
\captionsetup{justification=justified}

    \caption{Qualitative visualization of absolute segmentation error maps ($|GT - \text{Prediction}|$) on TG3K, TN3K, DDTI, and PKTN for ThyCLIPNet and eight ablated variants. Each row shows absolute error, highlighting boundary and shape discrepancies.}
    \label{pdf:ThyCLIPNet-diff}
\end{figure*}

\begin{figure*}[!ht]
\centering
\includegraphics[scale=0.63]{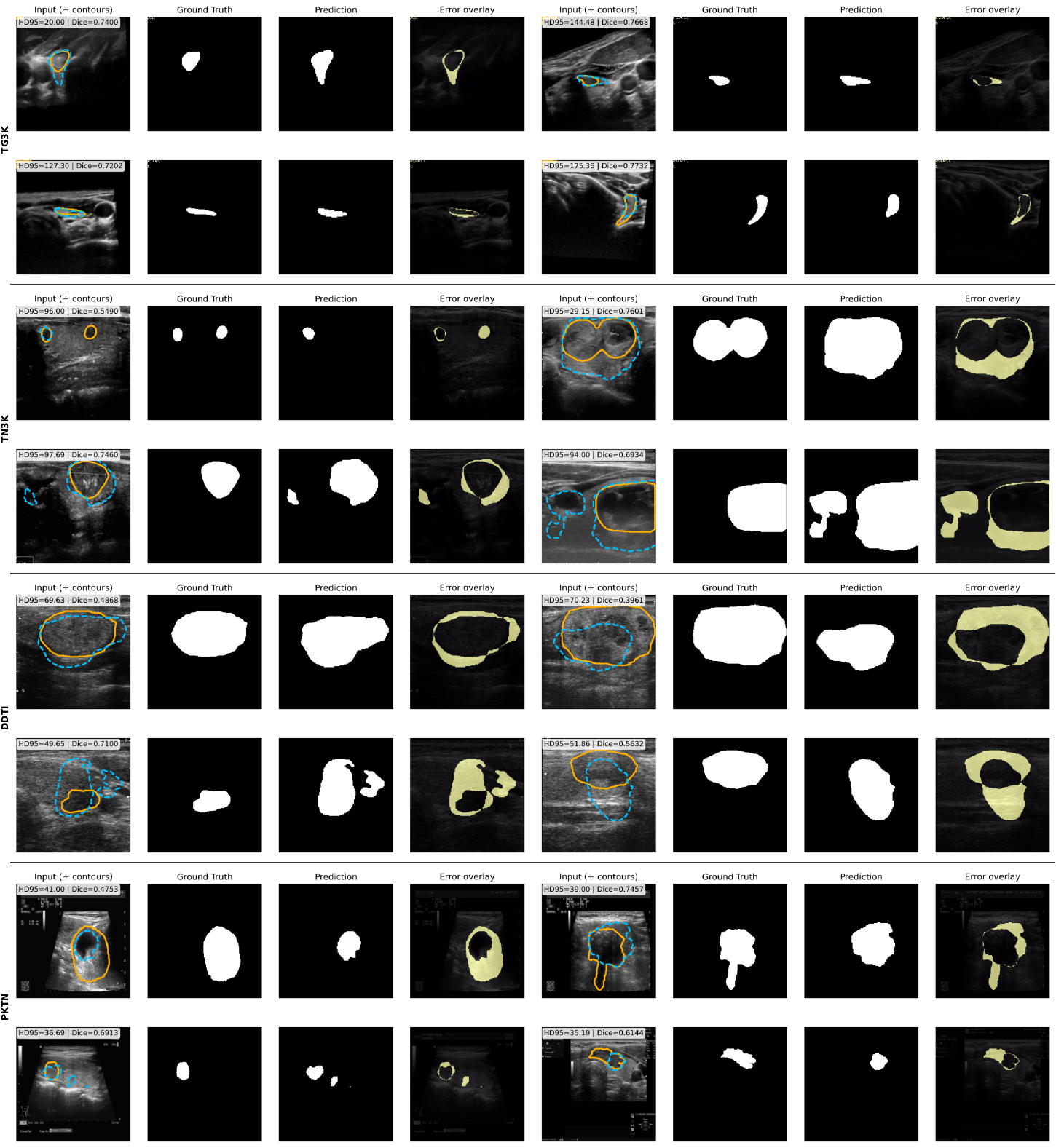}
\captionsetup{justification=justified}

\caption{ Representative hard-case segmentation results across four thyroid ultrasound datasets (TG3K, TN3K, DDTI, PKTN). Each row depicts a difficult example characterized by low contrast, boundary ambiguity, or structural noise. Columns show (from left to right): input image with annotated contours (ground truth in orange, prediction in blue), ground truth mask, predicted mask, and absolute error overlay. DICE and HD95 are reported per case to illustrate how boundary ambiguity elevates HD95 and how errors are typically concentrated near challenging edges.
}
\label{hard_cases}
\end{figure*}

To visually evaluate segmentation performance, \hyperref[pdf:ThyCLIPNet-diff]{Figure~\ref*{pdf:ThyCLIPNet-diff}} – \hyperref[hard_cases]{Figure~\ref*{hard_cases}} present representative qualitative results from all four thyroid ultrasound datasets (TG3K, TN3K, DDTI, and PKTN). \hyperref[pdf:ThyCLIPNet-diff]{Figure~\ref*{pdf:ThyCLIPNet-diff}} shows segmentation error maps ($|GT - \text{Prediction}|$) across nine model variants, including the proposed ThyCLIPNet and eight ablation configurations (without ECA, ASPP, BiomedCLIP, custom CBAM, lightweight CBAM, both CBAM variants, BiomedCLIP+ECA, and BiomedCLIP+ASPP). Error intensities highlight spatial discrepancies between predictions and ground truth, particularly along gland boundaries. Among all variants, ThyCLIPNet consistently produces the smallest error regions, confirming the additive benefits of its attention mechanisms and multi-scale fusion design. These visual results complement the quantitative findings, emphasizing the model’s robustness under diverse ultrasound conditions.

\begin{figure*}[!ht]
\centering
\includegraphics[scale=0.285]{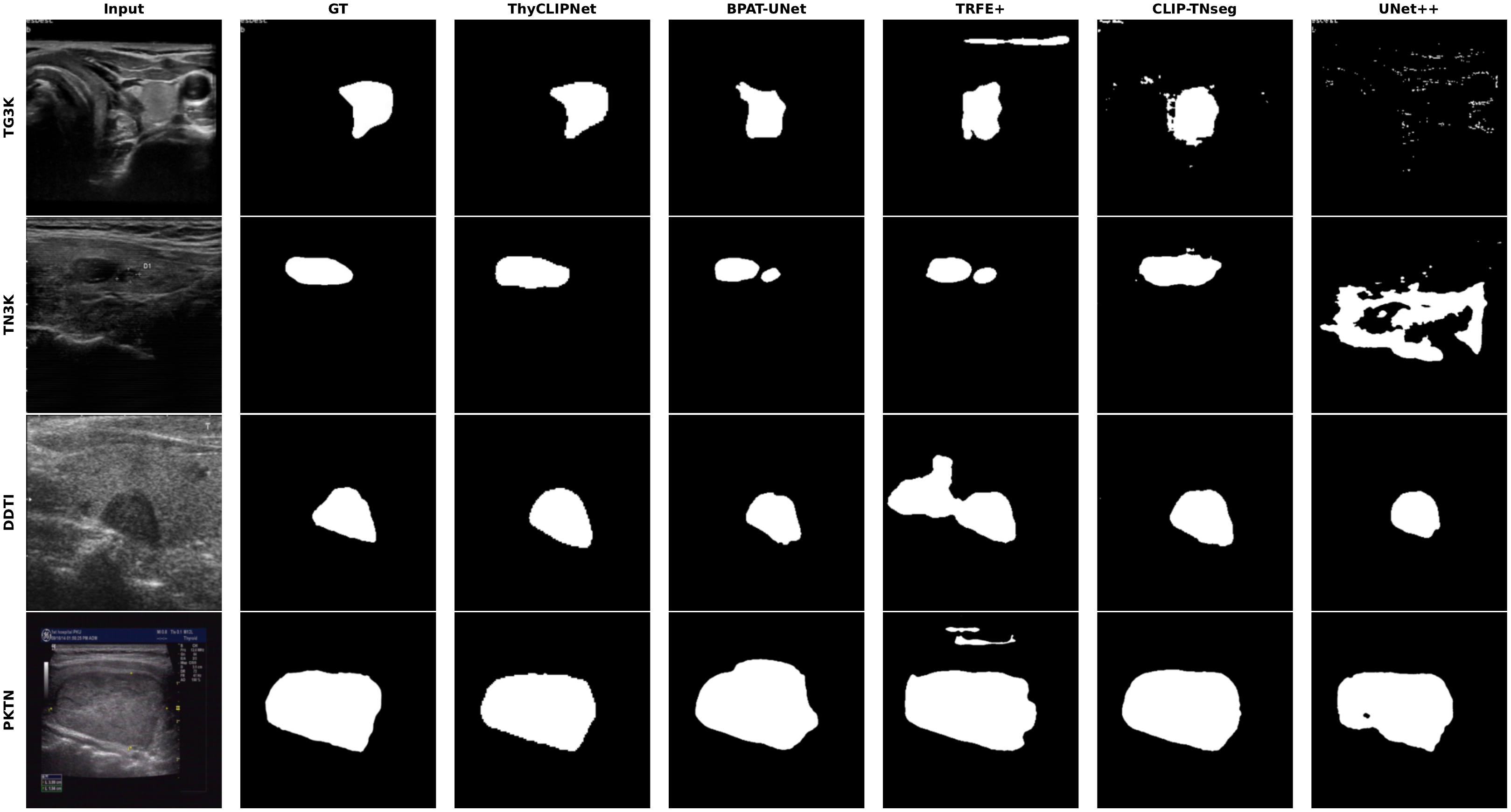}
\captionsetup{justification=justified}

\caption{Qualitative prediction matrix on representative test images from TG3K, TN3K, DDTI, and PKTN. Rows correspond to datasets; columns show the input ultrasound image, ground truth (GT), and predicted masks from ThyCLIPNet, BPAT-UNet, TRFE++, CLIP-TNseg, and UNet++.}
\label{fig:image_matrix}
\end{figure*}

Beyond the ablation study, \hyperref[fig:image_matrix]{Figure~\ref*{fig:image_matrix}} presents a qualitative comparison between ThyCLIPNet and representative state-of-the-art (SOTA) architectures: BPAT-UNet, TRFE+, CLIP-TNseg, and UNet++, using one representative case from each dataset. Compared to these baselines, ThyCLIPNet produces contiguous and morphologically consistent gland masks with minimal false positives and negligible boundary drift. In contrast, competing methods exhibit distinct artifacts, including fragmentation (e.g., broken glands on TN3K), spurious activations (e.g., scattered pixels on TG3K), and shape distortions (e.g., over-extended or under-segmented boundaries on DDTI and PKTN). These results visually demonstrate ThyCLIPNet’s enhanced stability, shape fidelity, and suppression of background noise.

To further evaluate robustness under challenging imaging conditions, \hyperref[hard_cases]{Figure~\ref*{hard_cases}} shows segmentation outputs on a curated set of complex cases characterized by low tissue contrast, boundary ambiguity, and irregular gland morphology. These factors are often responsible for elevated HD95 values. Across these cases, ThyCLIPNet maintains accurate contour alignment and stable region coverage, as evidenced by consistent DICE scores and reduced HD95. The visual overlays indicate that even in highly degraded or noisy scans, ThyCLIPNet preserves gland continuity and effectively suppresses spurious boundary deviations. 

\subsection{Ablation Study}
To explore the contribution of each architectural component in ThyCLIPNet, we performed an ablation study across all four datasets (TG3K, TN3K, DDTI, and PKTN). The full model was compared against eight ablated variants, each omitting specific modules: ECA, ASPP, BiomedCLIP, custom CBAM, lightweight CBAM, both CBAM variants, BiomedCLIP+ECA, and BiomedCLIP+ASPP depicted in \hyperref[tab:ablation_study]{Table~\ref*{tab:ablation_study}}.

\begin{table*}[ht]
\captionsetup{labelformat=empty, justification=raggedright, singlelinecheck=false}
\caption{\textbf{Table~\ref*{tab:ablation_study}.} Ablation‐study results on four datasets (TG3K, TN3K, DDTI, and PKTN). Metrics are mean\,$\pm$\, standard deviation over five independent runs. DICE, IoU, accuracy, recall, precision and HD95 are shown.}
\label{tab:ablation_study}
\centering
\resizebox{\textwidth}{!}{%
  \begin{tabular}{l l c c c c c c}
    \hline
    \textbf{Dataset} & \textbf{Architecture}
      & \textbf{DICE (\%) $\uparrow$}    & \textbf{IoU (\%) $\uparrow$}     & \textbf{Accuracy (\%) $\uparrow$} & \textbf{Recall (\%) $\uparrow$}   & \textbf{Precision (\%) $\uparrow$}  & \textbf{HD95 (px) $\downarrow$}      \\
    \hline
    
    \multirow{9}{*}{TG3K \cite{wunderling2017comparison}}
     & ThyCLIPNet w/o ECA                 
       & 96.08\,$\pm$\,0.10 & 92.45\,$\pm$\,0.20 & 99.31\,$\pm$\,0.00 & 97.07\,$\pm$\,0.20 & 95.11\,$\pm$\,0.30 & \cellcolor{bestcolor}\textbf{3.63\,$\pm$\,0.46} \\
     & ThyCLIPNet w/o ASPP                
       & 96.12\,$\pm$\,0.20 & 92.53\,$\pm$\,0.40 & 99.32\,$\pm$\,0.00 & 96.87\,$\pm$\,0.50 & 95.38\,$\pm$\,0.50 & 3.96\,$\pm$\,0.23 \\
     & ThyCLIPNet w/o BiomedCLIP          
       & 94.21\,$\pm$\,0.20 & 89.06\,$\pm$\,0.40 & 98.94\,$\pm$\,0.00 & 98.34\,$\pm$\,0.40 & 90.43\,$\pm$\,0.60 & 4.94\,$\pm$\,0.42 \\
     & ThyCLIPNet w/o custom CBAM         
       & 95.91\,$\pm$\,0.60 & 92.15\,$\pm$\,1.10 & 99.28\,$\pm$\,0.10 & 96.80\,$\pm$\,0.70 & 95.04\,$\pm$\,0.80 & 4.02\,$\pm$\,0.19 \\
     & ThyCLIPNet w/o lightweight CBAM    
       & 96.36\,$\pm$\,0.10 & 92.97\,$\pm$\,0.20 & 99.36\,$\pm$\,0.00 & 96.95\,$\pm$\,0.10 & 95.78\,$\pm$\,0.20 & 3.79\,$\pm$\,0.23 \\
     & ThyCLIPNet w/o both CBAM variants       
       & \cellcolor{bestcolor}\textbf{96.54\,$\pm$\,0.20} & \cellcolor{bestcolor}\textbf{93.32\,$\pm$\,0.30} & \cellcolor{bestcolor}\textbf{99.39\,$\pm$\,0.00}
         & 97.18\,$\pm$\,0.10 & \cellcolor{bestcolor}\textbf{95.92\,$\pm$\,0.20} & 3.75\,$\pm$\,0.13 \\
     & ThyCLIPNet w/o (BiomedCLIP+ECA)   
       & 96.11\,$\pm$\,0.10 & 92.51\,$\pm$\,0.20 & 99.30\,$\pm$\,0.00 & \cellcolor{bestcolor}\textbf{98.68\,$\pm$\,0.10} & 93.66\,$\pm$\,0.30 & 4.05\,$\pm$\,0.15 \\
     & ThyCLIPNet w/o (BiomedCLIP+ASPP)  
       & 95.74\,$\pm$\,0.30 & 91.82\,$\pm$\,0.60 & 99.23\,$\pm$\,0.10 & 98.60\,$\pm$\,0.10 & 93.04\,$\pm$\,0.60 & 4.15\,$\pm$\,0.23 \\
       \rowcolor{proposedcolor}
       & ThyCLIPNet-Proposed               
       & 96.22\,$\pm$\,0.10 & 92.72\,$\pm$\,0.10 & 99.33\,$\pm$\,0.00 & 96.80\,$\pm$\,0.20 & 95.64\,$\pm$\,0.20 & 3.75\,$\pm$\,0.16 \\
    \hline

    \multirow{9}{*}{TN3K \cite{gong2023thyroid}}
     & ThyCLIPNet w/o ECA                 
       & 87.37\,$\pm$\,0.60 & 77.57\,$\pm$\,0.90 & 96.85\,$\pm$\,0.20 & 90.16\,$\pm$\,0.80 & 84.77\,$\pm$\,1.60 & 16.43\,$\pm$\,1.09 \\
     & ThyCLIPNet w/o ASPP                
       & 87.07\,$\pm$\,0.30 & 77.11\,$\pm$\,0.50 & 96.77\,$\pm$\,0.10 & 90.15\,$\pm$\,0.50 & 84.19\,$\pm$\,0.30 & 16.91\,$\pm$\,0.88 \\
     & ThyCLIPNet w/o BiomedCLIP          
       & 86.56\,$\pm$\,0.60 & 76.30\,$\pm$\,1.00 & 96.54\,$\pm$\,0.20 & \cellcolor{bestcolor}\textbf{92.33\,$\pm$\,0.70} & 81.48\,$\pm$\,1.60 & 19.06\,$\pm$\,1.17 \\
     & ThyCLIPNet w/o custom CBAM         
       & 87.29\,$\pm$\,0.70 & 77.45\,$\pm$\,1.10 & 96.81\,$\pm$\,0.20 & 90.60\,$\pm$\,1.60 & 84.28\,$\pm$\,2.50 & 17.13\,$\pm$\,1.40 \\
     & ThyCLIPNet w/o lightweight CBAM    
       & 87.46\,$\pm$\,0.20 & 77.71\,$\pm$\,0.30 & 96.89\,$\pm$\,0.10 & 89.91\,$\pm$\,1.10 & 85.16\,$\pm$\,1.10 & \cellcolor{bestcolor}\textbf{16.21\,$\pm$\,0.62} \\
     & ThyCLIPNet w/o both CBAM variants       
       & 87.31\,$\pm$\,0.40 & 77.48\,$\pm$\,0.60 & 96.88\,$\pm$\,0.10 & 88.78\,$\pm$\,1.00 & \cellcolor{bestcolor}\textbf{85.91\,$\pm$\,1.40} & 16.32\,$\pm$\,0.46 \\
     & ThyCLIPNet w/o (BiomedCLIP+ECA)   
       & \cellcolor{bestcolor}\textbf{87.67\,$\pm$\,0.20} & \cellcolor{bestcolor}\textbf{78.04\,$\pm$\,0.30} & 96.91\,$\pm$\,0.10 & 91.05\,$\pm$\,0.90 & 84.54\,$\pm$\,1.10 & 16.79\,$\pm$\,1.22 \\
     & ThyCLIPNet w/o (BiomedCLIP+ASPP)  
       & 87.10\,$\pm$\,0.40 & 77.15\,$\pm$\,0.60 & 96.75\,$\pm$\,0.20 & 90.92\,$\pm$\,1.50 & 83.63\,$\pm$\,1.80 & 17.65\,$\pm$\,1.48 \\
        \rowcolor{proposedcolor}
       & ThyCLIPNet-Proposed               
       & 87.58\,$\pm$\,0.30 & 77.91\,$\pm$\,0.50 & \cellcolor{bestcolor}\textbf{96.94\,$\pm$\,0.10} & 89.47\,$\pm$\,0.40 & 85.78\,$\pm$\,0.50 & 16.38\,$\pm$\,0.92 \\
    \hline

    \multirow{9}{*}{DDTI \cite{pedraza2015open}}
     & ThyCLIPNet w/o ECA                 
       & 84.08\,$\pm$\,0.40 & 72.53\,$\pm$\,0.60 & 94.98\,$\pm$\,0.10 & 90.00\,$\pm$\,1.00 & 78.89\,$\pm$\,0.70 & 18.45\,$\pm$\,0.83 \\
     & ThyCLIPNet w/o ASPP                
       & 84.23\,$\pm$\,1.00 & 72.76\,$\pm$\,1.50 & 95.08\,$\pm$\,0.40 & 89.07\,$\pm$\,1.40 & 79.94\,$\pm$\,2.30 & 18.73\,$\pm$\,1.84 \\
     & ThyCLIPNet w/o BiomedCLIP          
       & 79.33\,$\pm$\,1.40 & 65.76\,$\pm$\,2.00 & 92.97\,$\pm$\,0.70 & \cellcolor{bestcolor}\textbf{91.45\,$\pm$\,3.00} & 70.15\,$\pm$\,2.80 & 26.91\,$\pm$\,2.92 \\
     & ThyCLIPNet w/o custom CBAM         
       & 84.23\,$\pm$\,1.10 & 72.77\,$\pm$\,1.70 & 95.05\,$\pm$\,0.50 & 89.48\,$\pm$\,1.50 & 79.63\,$\pm$\,2.70 & 18.51\,$\pm$\,1.44 \\
     & ThyCLIPNet w/o lightweight CBAM    
       & 84.07\,$\pm$\,1.70 & 72.55\,$\pm$\,2.50 & 95.11\,$\pm$\,0.60 & 87.35\,$\pm$\,1.40 & \cellcolor{bestcolor}\textbf{81.07\,$\pm$\,2.80} & 18.44\,$\pm$\,1.62 \\
     & ThyCLIPNet w/o both CBAM variants       
       & 83.74\,$\pm$\,0.70 & 72.04\,$\pm$\,1.10 & 94.94\,$\pm$\,0.40 & 88.30\,$\pm$\,2.00 & 79.73\,$\pm$\,2.80 & 19.19\,$\pm$\,2.68 \\
     & ThyCLIPNet w/o (BiomedCLIP+ECA)   
       & 82.27\,$\pm$\,1.10 & 69.89\,$\pm$\,1.60 & 94.42\,$\pm$\,0.30 & 87.87\,$\pm$\,2.30 & 77.37\,$\pm$\,1.30 & 20.78\,$\pm$\,1.68 \\
     & ThyCLIPNet w/o (BiomedCLIP+ASPP)  
       & 80.33\,$\pm$\,1.30 & 67.14\,$\pm$\,1.80 & 93.71\,$\pm$\,0.40 & 87.23\,$\pm$\,3.00 & 74.49\,$\pm$\,1.20 & 22.43\,$\pm$\,1.52 \\
        \rowcolor{proposedcolor}
        & ThyCLIPNet-Proposed               
       & \cellcolor{bestcolor}\textbf{84.73\,$\pm$\,0.50} & \cellcolor{bestcolor}\textbf{73.51\,$\pm$\,0.70} & \cellcolor{bestcolor}\textbf{95.26\,$\pm$\,0.20} & 89.25\,$\pm$\,1.70 & 80.69\,$\pm$\,1.80 & \cellcolor{bestcolor}\textbf{18.23\,$\pm$\,1.10} \\
    \hline
    
    \multirow{9}{*}{PKTN \cite{sun2025clip}}
     & ThyCLIPNet w/o ECA                 
       & \cellcolor{bestcolor}\textbf{80.91\,$\pm$\,0.50} & \cellcolor{bestcolor}\textbf{67.94\,$\pm$\,0.70} & \cellcolor{bestcolor}\textbf{98.70\,$\pm$\,0.00} & 80.00\,$\pm$\,2.00 & 81.90\,$\pm$\,1.30 & 12.65\,$\pm$\,1.71 \\
     & ThyCLIPNet w/o ASPP                
       & 79.97\,$\pm$\,0.90 & 66.64\,$\pm$\,1.30 & 98.65\,$\pm$\,0.10 & 78.15\,$\pm$\,1.50 & 81.92\,$\pm$\,1.80 & 11.40\,$\pm$\,1.12 \\
     & ThyCLIPNet w/o BiomedCLIP          
       & 80.19\,$\pm$\,1.00 & 66.94\,$\pm$\,1.30 & 98.62\,$\pm$\,0.10 & 80.83\,$\pm$\,0.30 & 79.58\,$\pm$\,1.80 & 13.78\,$\pm$\,2.81 \\
     & ThyCLIPNet w/o custom CBAM         
       & 80.66\,$\pm$\,0.90 & 67.60\,$\pm$\,1.30 & 98.69\,$\pm$\,0.10 & 78.85\,$\pm$\,1.80 & \cellcolor{bestcolor}\textbf{82.60\,$\pm$\,1.40} & 11.74\,$\pm$\,1.39 \\
     & ThyCLIPNet w/o lightweight CBAM    
       & 80.55\,$\pm$\,1.10 & 67.44\,$\pm$\,1.50 & 98.69\,$\pm$\,0.10 & 78.79\,$\pm$\,2.30 & 82.44\,$\pm$\,1.60 & 11.15\,$\pm$\,1.10 \\
     & ThyCLIPNet w/o both CBAM variants       
       & 80.63\,$\pm$\,1.10 & 67.56\,$\pm$\,1.60 & 98.69\,$\pm$\,0.10 & 78.91\,$\pm$\,2.20 & 82.47\,$\pm$\,1.10 & \cellcolor{bestcolor}\textbf{10.16\,$\pm$\,0.61} \\
     & ThyCLIPNet w/o (BiomedCLIP+ECA)   
       & 80.27\,$\pm$\,0.90 & 67.05\,$\pm$\,1.20 & 98.59\,$\pm$\,0.10 & 82.69\,$\pm$\,1.00 & 78.04\,$\pm$\,2.40 & 15.19\,$\pm$\,3.58 \\
     & ThyCLIPNet w/o (BiomedCLIP+ASPP)  
       & 79.60\,$\pm$\,0.70 & 66.12\,$\pm$\,1.00 & 98.53\,$\pm$\,0.10 & \cellcolor{bestcolor}\textbf{82.76\,$\pm$\,1.20} & 76.69\,$\pm$\,1.10 & 15.22\,$\pm$\,1.23 \\
       \rowcolor{proposedcolor}
       & ThyCLIPNet-Proposed               
       & 80.70\,$\pm$\,0.40 & 67.64\,$\pm$\,0.50 & 98.67\,$\pm$\,0.00 & 80.22\,$\pm$\,1.80 & 81.25\,$\pm$\,2.00 & 10.86\,$\pm$\,0.46 \\
    \hline
  \end{tabular}%
}
\vspace{0.2cm}
\parbox{\textwidth}{%
  \footnotesize
  \textbf{Color Legend:} \colorbox{proposedcolor}{\phantom{XX}} Proposed method (ThyCLIPNet), 
  \colorbox{bestcolor}{\phantom{XX}} Best performance per metric (including ablation variants); \textbf{Bold}: Best performance per metric per dataset; ($\uparrow$: higher is better, $\downarrow$: lower is better).
}
\end{table*}

Since the four datasets differ substantially in anatomical complexity, noise characteristics, and boundary ambiguity, we do not define the optimal architecture based on isolated improvements on a single metric or dataset. Instead, we adopt the following selection criteria: (i) cross-dataset stability, (ii) strong overlap performance (DICE/IoU), (iii) preserved boundary fidelity (HD95), which reflects contour reliability, (iv) balanced precision–recall behaviour to avoid over-segmentation, and (v) a favorable efficiency–accuracy trade-off. Under this multi-objective criterion, the complete ThyCLIPNet provides the most reliable and practically balanced configuration. While some ablations yield localized gains on a single dataset or metric, they do not generalize and typically trade off HD95 or precision, indicating dataset-specific sensitivity rather than systematic improvement.

Across the four datasets, the proposed model achieves the highest average DICE ($\approx 87.31\%$) among all variants, indicating the strongest overall overlap consistency. Although certain ablated variants achieve marginal DICE improvements on individual datasets (e.g., TG3K or TN3K), these gains are small in magnitude and, in several cases, comparable to the observed standard deviation range. More importantly, such localized improvements do not generalize across datasets. For example, the CBAM-free variant attains the highest DICE on TG3K (96.54\%), yet its DICE drops on DDTI (83.74\%) and does not improve PKTN performance (80.63\%). This indicates dataset-specific sensitivity rather than systematic architectural superiority.

The contribution of BiomedCLIP is most evident on challenging datasets. On DDTI, removing BiomedCLIP produces the largest degradations: DICE decreases from 84.73 to 79.33 (-5.40 pp), IoU decreases from 73.51 to 65.76 (-7.75 pp), and HD95 worsens from 18.23 to 26.91 (+8.68). Notably, although recall increases (89.25 to 91.45\%), precision drops sharply (80.69 to 70.15\%), which is consistent with over-segmentation rather than improved delineation. DDTI also exhibits the largest performance variance across ablation variants, making it a discriminative benchmark for assessing architectural sensitivity across variants. A similar behavior is observed on TN3K: recall rises when BiomedCLIP is removed (89.47 to 92.33\%), but DICE decreases (87.58 to 86.56\%), precision declines (85.78 to 81.48\%), and HD95 worsens (16.38 to 19.06). Since boundary accuracy (HD95) is particularly important for reliable nodule measurement in thyroid assessment, the proposed configuration prioritizes anatomically consistent delineation over apparent recall gains caused by over-segmentation.

To quantitatively assess cross-dataset consistency, we compute a cross-dataset mean-rank score (Supplementary Table S3). Under this analysis, ThyCLIPNet achieves the lowest overall mean rank (2.3125), indicating the most consistent multi-metric performance across datasets. This provides a descriptive summary of cross-dataset multi-metric consistency of the proposed configuration.

Later on, we assessed the efficiency trade-offs of those variants that showed accuracy gains by comparing their parameter counts, total training time, and per-image inference latency on the TG3K dataset (\hyperref[tab:efficiency]{Table~\ref{tab:efficiency}}). We selected TG3K as our efficiency benchmark because it has the greatest computational demand, with the longest total training time and highest step count, making efficiency differences more observable. 
To quantify the computational efficiency of each model variant, we computed three primary metrics: params (number of trainable weights), train time (total time to complete all training epochs), and inf time (average per-image inference latency).

The corresponding $\Delta$ columns ($\Delta$Params, $\Delta$Train, $\Delta$Inf) show the percentage change of each metric relative to the proposed ThyCLIPNet baseline, which is fixed at 0 in all $\Delta$ columns. The values are computed as:

\begin{equation}
\Delta X = \left( \frac{X_{\text{variant}} - X_{\text{baseline}}}{X_{\text{baseline}}} \right) \times 100
\end{equation}
Here, negative $\Delta$ values denote improvements (smaller size or faster execution), while positive values indicate regressions.

Removing ECA incurs a 5.93\% increase in total training time while slightly increasing inference latency by 2.84\%. Omitting both Biomed-CLIP and ECA slows training by 19.51\% and also substantially increases inference time. While dropping BiomedCLIP alone reduces training time by 44.11\%, it produces severe segmentation losses (up to 5.4 pp DICE, 7.75 pp IoU) on the DDTI dataset, despite a modest inference reduction (0.01148s, -9.54\%). The smallest variant, w/o (BiomedCLIP+ASPP), reduces parameters by 46.86\% and achieves the largest inference reduction (0.00879 s, -30.73\%); however, it still requires 14,859.38 s to train and degrades segmentation performance (\hyperref[tab:ablation_study]{Table~\ref*{tab:ablation_study}}). Meanwhile, the CBAM-free variant delivers the highest DICE and IoU on the TG3K dataset but yields only a 3.28\% parameter reduction, increases training time by 32.12\%, and offers a negligible inference-time change (0.01267 s, -0.16\%). A similar trend is observed without the lightweight CBAM variant, which shows slightly improved DICE and IoU on the TG3K dataset, reduces training time by 0.58\% and also reduces inference time (0.01150s, –9.38\%), but increases HD95 and has no consistent advantage across other datasets. These findings reinforce that the proposed ThyCLIPNet model (8.55M parameters, 15,170.77s training, 0.01269s per-image inference) remains the most balanced choice in terms of segmentation quality, model size, and computational efficiency.

\begin{table*}[ht]
\captionsetup{labelformat=empty, justification=raggedright, singlelinecheck=false}
\caption{\textbf{Table~\ref*{tab:efficiency}.} Efficiency comparison on TG3K. Values are reported as the mean over five independent runs. $\Delta$ values are relative to the proposed ThyCLIPNet model.}

\label{tab:efficiency}
\centering
\resizebox{\textwidth}{!}{%
  \begin{tabular}{l c c c c c c}
    \hline
    \textbf{Variants}                         & \textbf{Params}     & \textbf{$\Delta$Params (\%)} & \textbf{Train Time (s)} & \textbf{$\Delta$Train (\%)} & \textbf{Inf Time (s)} & \textbf{$\Delta$Inf (\%)} \\
    \hline
    w/o ECA                          & 8,550,704  & \cellcolor{improvecolor}0.00        & {16,070.82} & \cellcolor{regresscolor}{+5.93}  & {0.01305} & \cellcolor{regresscolor}{+2.84}  \\    
    w/o BiomedCLIP                   & 7,821,736  & \cellcolor{improvecolor}-8.53       &  {8,479.56} & \cellcolor{improvecolor}{-44.11} & {0.01148} & \cellcolor{improvecolor}{-9.54}  \\
    w/o lightweight CBAM             & 8,312,114  & \cellcolor{improvecolor}-2.79       & {15,083.26} & \cellcolor{improvecolor}{-0.58}  & {0.01150} & \cellcolor{improvecolor}{-9.38}  \\
    w/o both CBAM variants           & 8,270,353  & \cellcolor{improvecolor}-3.28       & {20,043.51} & \cellcolor{regresscolor}{+32.12} & {0.01267} & \cellcolor{improvecolor}{-0.16}  \\
    w/o (BiomedCLIP+ECA)             & 7,821,720  & \cellcolor{improvecolor}-8.53       & {18,130.88} & \cellcolor{regresscolor}{+19.51} & {0.01874} & \cellcolor{regresscolor}{+47.68} \\
    w/o (BiomedCLIP+ASPP)            & 4,543,656  & \cellcolor{improvecolor}-46.86      & {14,859.38} & \cellcolor{improvecolor}{-2.05}  & {0.00879} & \cellcolor{improvecolor}{-30.73} \\ 
    \textbf{ThyCLIPNet-Proposed}     & 8,550,720  & 0.00                               & {15,170.77} & {0.00}                          & {0.01269} & {0.00} \\
    \hline
  \end{tabular}%
}
\vspace{0.2cm}
\parbox{\textwidth}{%
  \footnotesize
  \textbf{Color Legend:} \colorbox{improvecolor}{\phantom{XX}} Improvement (beneficial change), \colorbox{regresscolor}{\phantom{XX}} Regression (detrimental change).
}
\end{table*}

Taken together, the ablation and efficiency analyses demonstrate that simplified variants either produce localized metric gains without cross-dataset stability or reduce computational cost at the expense of structural segmentation quality, particularly boundary fidelity (HD95). No ablated configuration simultaneously improves overlap metrics, preserves boundary fidelity, and maintains efficiency. Therefore, the proposed ThyCLIPNet configuration is selected not for isolated metric maximization but for the most robust multi-objective trade-off across datasets.

Post-processing with test-time augmentation (TTA) further enhances the robustness of our proposed model. As demonstrated in \hyperref[tab:tta]{Table~\ref*{tab:tta}}, enabling TTA consistently improves DICE, IoU, accuracy, and recall across the four datasets. Although these gains may seem modest, they indicate that the model benefits from inference-time ensembling, particularly when dealing with anatomically diverse or noisy images.

\begin{table*}
\captionsetup{labelformat=empty, justification=raggedright, singlelinecheck=false}
\centering
\caption{\textbf{Table~\ref*{tab:tta}.} Performance of ThyCLIPNet + TTA across four datasets metrics as mean\,$\pm$\,std.}
\label{tab:tta}
\resizebox{\textwidth}{!}{%
  \begin{tabular}{l l c c c c c c}
    \hline
    \textbf{Dataset} & \textbf{Architecture}
      & \textbf{DICE (\%)}    & \textbf{IoU (\%)}      & \textbf{Accuracy (\%)} & \textbf{Recall (\%)}   & \textbf{Precision (\%)} & \textbf{HD95 (px)}      \\
    \hline
    \rowcolor{ttacolor}
    TG3K & ThyCLIPNet+TTA
      & \textbf{96.43\,$\pm$\,0.10} & \textbf{93.10\,$\pm$\,0.10} & \textbf{99.37\,$\pm$\,0.00} & \textbf{97.03\,$\pm$\,0.20} & \textbf{95.83\,$\pm$\,0.20} &  \textbf{3.69\,$\pm$\,0.16} \\
    \rowcolor{ttacolor}
    TN3K & ThyCLIPNet+TTA
      & \textbf{87.92\,$\pm$\,0.20} & \textbf{78.45\,$\pm$\,0.30} & \textbf{97.03\,$\pm$\,0.10} & 89.49\,$\pm$\,0.70 & \textbf{86.42\,$\pm$\,0.80} & \textbf{16.09\,$\pm$\,0.88} \\
    \rowcolor{ttacolor}
    DDTI & ThyCLIPNet+TTA
      & \textbf{85.05\,$\pm$\,0.50} & \textbf{73.99\,$\pm$\,0.80} & \textbf{95.35\,$\pm$\,0.20} & \textbf{89.72\,$\pm$\,1.40} & 80.88\,$\pm$\,1.80 & 18.41\,$\pm$\,1.28 \\
    \rowcolor{ttacolor}
    PKTN & ThyCLIPNet+TTA
      & \textbf{80.77\,$\pm$\,0.30} & \textbf{67.75\,$\pm$\,0.40} & \textbf{98.68\,$\pm$\,0.00} & \textbf{80.45\,$\pm$\,1.80} & 81.17\,$\pm$\,2.00 & 10.91\,$\pm$\,0.51 \\
    \hline
  \end{tabular}%
}
\vspace{0.1cm}
\parbox{\textwidth}{%
  \footnotesize
  \textbf{Color Legend:} \colorbox{ttacolor}{\phantom{XX}} ThyCLIPNet with Test Time Augmentation (TTA); 
  \textbf{Bold}: Metric improved by TTA compared to base model
}
\end{table*}

\subsection{Computational Efficiency and Complexity Analysis}
To strengthen the analysis of model complexity and computational efficiency, we conducted a detailed comparison of the proposed ThyCLIPNet with both reimplemented baselines and previously reported thyroid segmentation networks. 
The evaluation considers parameter count, FLOPs, inference latency, and GPU memory usage, providing a transparent assessment of ThyCLIPNet's efficiency relative to established architectures. In the adopted workflow, the frozen BiomedCLIP vision encoder is used for one-time feature extraction, with 512-dimensional embeddings precomputed and stored before subsequent segmentation training and inference.

As shown in \hyperref[tab:efficiency_analysis]{Table~\ref*{tab:efficiency_analysis}}, ThyCLIPNet provides a strong accuracy-efficiency trade-off among the compared architectures. For the representative baselines included in this table, inference-time benchmarking was conducted under identical runtime settings, while representative lightweight baselines were additionally re-implemented and evaluated under the same preprocessing and inference protocol as ThyCLIPNet. Although it is not the smallest model, the ThyCLIPNet segmentation stage remains compact (8.55M parameters) and computationally efficient (22.99 G FLOPs), while maintaining fast inference (0.01269 s/img; 78.81 FPS). Compared to UNet++ \cite{zhou2018unet++}, ThyCLIPNet reduces computational load by approximately 67\% (69.43 G vs. 22.99 G) and improves inference performance from 73.58 to 78.81 FPS. Against more complex transformer-based and dual-decoder architectures like TRFE+ \cite{GONG2023106389} and BPAT-UNet \cite{BI2023107614}, ThyCLIPNet is approximately 3$\times$ and 6$\times$ faster, respectively, while being $\sim 5\times$ and $\sim 8\times$ smaller in parameter count, respectively, yet maintaining competitive segmentation performance.

When compared with representative lightweight baselines, including DeepLabV3+ with MobileNetV2 \cite{chen2018encoder} (5.85 M params; 0.01223 s/img) and U-Net with MobileNetV2 \cite{ronneberger2015u} (3.75 M params; 0.00698 s/img), these baselines exhibit lower parameter counts and, in some cases, faster raw inference. However, they show consistently weaker overlap performance than ThyCLIPNet on TN3K and DDTI. In addition, on TN3K, ThyCLIPNet demonstrates improved boundary accuracy relative to both lightweight baselines. These results indicate that ThyCLIPNet achieves a favorable balance between segmentation performance and computational efficiency while maintaining efficient inference through BiomedCLIP-guided
semantic fusion and attention refinement.


\begin{table*}[ht]
\centering
\scriptsize
\setlength{\tabcolsep}{6pt}
\renewcommand{\arraystretch}{1.1}
\captionsetup{labelformat=empty, justification=justified, singlelinecheck=false}
\caption{\textbf{Table~\ref*{tab:efficiency_analysis}.} Computational benchmark comparison of ThyCLIPNet with representative baselines. Representative lightweight models were re-implemented under identical evaluation settings (input size = 256 × 256, batch size = 1), while remaining architectures were benchmarked under the same hardware and inference environment. FLOPs and parameter counts are architecture-dependent and dataset-independent. Inference times represent median (P50) values over five independent runs; for ThyCLIPNet, computational measurements correspond to the segmentation stage using precomputed BiomedCLIP embeddings. FPS is computed as $1/\text{time}$.}

\label{tab:efficiency_analysis}

\resizebox{\textwidth}{!}{%
\begin{tabular}{lccccc}
    \hline
    \textbf{Model} & \textbf{Params (M)} & \textbf{Inference Time (s/img)} & \textbf{FPS} & \textbf{FLOPs (G)} & \textbf{GPU Mem (GB)} \\
    \hline
    UNet++\cite{zhou2018unet++} & {9.16} & {0.01359} & {73.58} & {69.43} & {1.81} \\
    TRFE+ \cite{GONG2023106389} & {45.44} & {0.03834} & {26.34} & {160.77} & {0.76} \\
    BPAT-UNet\cite{BI2023107614} & {72.39} & {0.07648} & {13.08} & {128.66} & {0.55} \\
    
    DeepLabV3+ with MobileNetV2\cite{chen2018encoder} & 5.85 & 0.01223 & 81.77 & \textbf{12.37} & 0.77 \\
    U-Net with MobileNetV2\cite{ronneberger2015u} & \textbf{3.75} & \textbf{0.00698} & \textbf{143.27} & 15.85 & \textbf{0.52} \\
    \rowcolor{proposedcolor}
    \textbf{ThyCLIPNet-Proposed} & {8.55} & {0.01269} & {78.81} & {22.99} & {0.76} \\
    \hline
\end{tabular}%
}
\vspace{0.15cm}
\parbox{\textwidth}{%
  \footnotesize
\textbf{Bold:} Best performance among compared models; 
  \colorbox{proposedcolor}{\phantom{XX}} Proposed model (ThyCLIPNet).
}
\end{table*}

Compared with previously published models in \hyperref[tab:reported_efficiency]{Table~\ref*{tab:reported_efficiency}}, ThyCLIPNet segmentation stage demonstrates favorable inference speed and memory efficiency under the adopted precomputation workflow. It achieves $\sim$6$\times$ higher FPS than TNSNet \cite{sun2022tnsnet} (12.5 FPS) and is $\sim$18$\times$ faster than MSAC-DD \cite{payatsuporn2025papillary} (0.233 s/img). Relative to FDE-Net \cite{chen2023fde} and DualSwinUNet++ \cite{dialameh2025dualswinunet++}, ThyCLIPNet achieves $\sim$3$\times$ and $\sim$14$\times$ faster inference, respectively. Although transformer-based models such as SAMTNU \cite{chen2025samtnu} possess high representational capacity (130.99 M parameters), they incur approximately 6$\times$ higher computational cost in FLOPs (147.38 G vs. 22.99 G), which can limit their feasibility for real-time clinical applications. Moreover, the ThyCLIPNet segmentation stage requires only 0.76 GB of GPU memory, substantially lower than TNSNet's \cite{sun2022tnsnet} 25.42 GB, reinforcing its suitability for edge devices and resource-constrained deployment. Overall, these results indicate that ThyCLIPNet maintains competitive segmentation accuracy while substantially reducing computational cost and memory usage compared with CNN and transformer baselines, supporting its potential suitability for real-time CAD pipelines and portable ultrasound systems under the adopted precomputation workflow. 
\begin{table*}[ht]
\centering
\scriptsize
\setlength{\tabcolsep}{6pt}
\renewcommand{\arraystretch}{1.1}
\captionsetup{labelformat=empty, justification=justified, singlelinecheck=false}
\caption{\textbf{Table~\ref*{tab:reported_efficiency}}. Comparison with previously reported thyroid segmentation networks. Values are primarily extracted from the corresponding original publications, whereas ThyCLIPNet was benchmarked under the experimental protocol adopted in this study (input size = 256 × 256, batch size = 1).}
 
\label{tab:reported_efficiency}

\resizebox{\textwidth}{!}{%
\begin{tabular}{lccccc}
    \hline
    \textbf{Model} & \textbf{Params (M)} & \textbf{Inference Time (s/img)} & \textbf{FPS} & \textbf{FLOPs (G)} & \textbf{GPU Mem (GB)} \\
    \hline
    GDSSA-Net~\cite{farooq2025gdssa} & {35.00} & {0.042} & \textemdash & \textemdash & \textemdash \\
    MSAC-DD~\cite{payatsuporn2025papillary} & {25.70} & {0.233} & \textemdash & \textemdash & {2.55} \\
    FCG-Net~\cite{shao2023fcg} & {11.35} & \textemdash & {20.25} & \textemdash & \textemdash \\
    FDE-Net~\cite{chen2023fde} & \textemdash & {0.036} & \textemdash & \textemdash & \textemdash \\
    DualSwinUNet++~\cite{dialameh2025dualswinunet++} & \textemdash & {0.180} & \textemdash & \textemdash & \textemdash \\
    TNSNet~\cite{sun2022tnsnet} & {46.47} & {0.080} & {12.5} & \textemdash & {25.42} \\
    SAMTNU~\cite{chen2025samtnu} & {130.99} & \textemdash & \textemdash & {147.38} & \textemdash \\
    \rowcolor{proposedcolor}
    \textbf{ThyCLIPNet-Proposed } & {\textbf{8.55}} & {\textbf{0.01269}} & {\textbf{78.81}} & {\textbf{22.99}} & {\textbf{0.76}} \\
    \hline
\end{tabular}%
}
\vspace{0.15cm}
\parbox{\textwidth}{%
  \footnotesize
  \textbf{Bold:} Best performance among compared models; 
  \colorbox{proposedcolor}{\phantom{XX}} Proposed model (ThyCLIPNet). 
  Missing entries (\textemdash) indicate values not reported in the original papers.
}
\end{table*}

\subsection{Grad-CAM Interpretability} 
\begin{figure*}[!ht]
\centerline{\includegraphics[scale=.30]{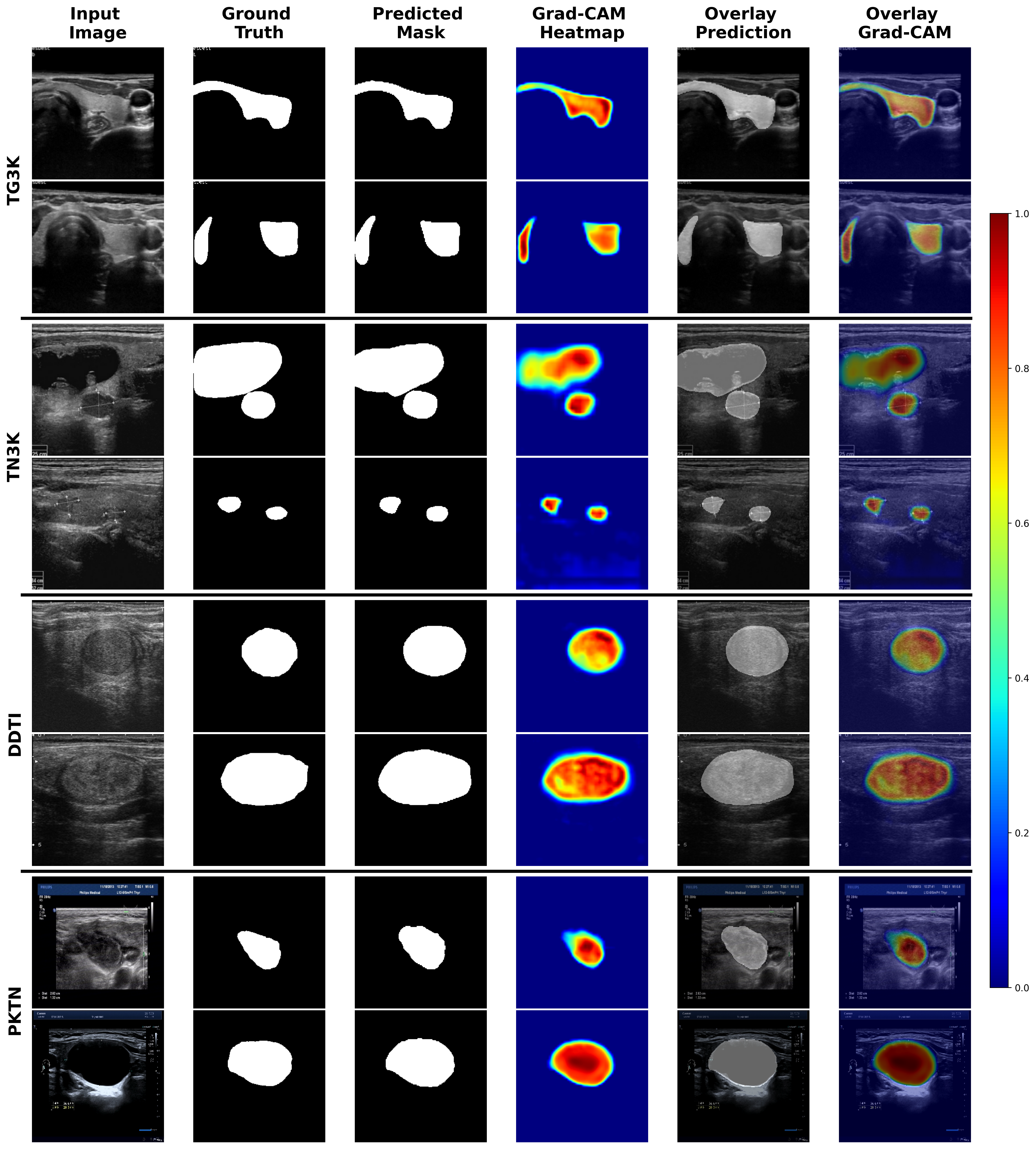}}
\captionsetup{justification=justified}

\caption{Six‐panel Grad-CAM interpretability visualizations for two representative cases from each thyroid ultrasound dataset (TG3K: top two rows; TN3K: next two rows; DDTI: next two rows; PKTN: bottom two rows). From left to right: (i) input image, (ii) ground-truth mask, (iii) predicted mask, (iv) Grad-CAM heatmap, (v) overlay of prediction, and (vi) overlay of Grad-CAM heatmap.}
\label{fig:gradcam}
\end{figure*}

To apply Grad-CAM in our dual-input segmentation model, we feed both the ultrasound image and the BiomedCLIP embedding to the network and compute gradients through a selected decoder convolutional layer. We define a scalar score as the sum of the foreground-class predictions and backpropagate its gradient to that layer. The gradients are spatially averaged to obtain per-channel weights; a heatmap is formed as the weighted sum of the feature maps, followed by ReLU, normalization to \([0,1]\), and bilinear resizing to the input resolution. Because the model’s forward pass already combines the ultrasound image with the BiomedCLIP embedding, the resulting maps reflect contributions from both inputs.
We apply this pipeline to two representative cases per dataset, yielding structured six-panel visualizations comprising: (i) Input image, (ii) Ground truth, (iii) Predicted mask, (iv) Grad-CAM heatmap, (v) Overlay prediction, and (vi) Overlay Grad-CAM, as illustrated in \hyperref[fig:gradcam]{Figure~\ref*{fig:gradcam}}.

\subsection{Discussion}
 The proposed ThyCLIPNet model is designed to balance segmentation performance with computational efficiency by combining lightweight local feature extraction with complementary global biomedical semantic guidance. The MobileNetV2-based encoder, supported by ECA and multiscale contextual aggregation at the bottleneck, preserves informative spatial representations without substantially increasing model complexity. During decoding, attention-based refinement and selective integration of BiomedCLIP-derived global visual embeddings at lower-resolution stages allow broader semantic information to complement locally preserved structural features, while high-resolution reconstruction remains primarily driven by spatial details from the CNN pathway and skip connections, as illustrated in \hyperref[fig:proposed-f]{Figure~\ref*{fig:proposed-f}}. Ablation studies show that the effects of individual components are metric- and dataset-dependent; however, the complete ThyCLIPNet configuration provides the most favorable aggregate cross-dataset performance and removal of BiomedCLIP notably reduces overlap quality and worsens HD95. Although some ablated variants achieve slight gains in individual metrics on particular datasets, these improvements are not consistently maintained across the evaluated datasets, as shown in \hyperref[tab:ablation_study]{Table~\ref*{tab:ablation_study}} and \hyperref[tab:efficiency]{Table~\ref*{tab:efficiency}}.  

The qualitative analyses, including error maps (\hyperref[pdf:ThyCLIPNet-diff]{Figure~\ref*{pdf:ThyCLIPNet-diff}}), prediction matrix (\hyperref[fig:image_matrix]{Figure~\ref*{fig:image_matrix}}) and hard-case visualizations (\hyperref[hard_cases]{Figure~\ref*{hard_cases}}), further reveal ThyCLIPNet’s robustness under adverse imaging conditions, such as blurred or low-contrast gland boundaries where competing models show notable errors. These visual results suggest that remaining errors are frequently concentrated around uncertain or poorly defined boundaries, consistent with the boundary-sensitive behavior reflected by HD95. Grad-CAM visualizations (\hyperref[fig:gradcam]{Figure~\ref*{fig:gradcam}}) further illustrate the spatial regions emphasized by the model during prediction. Alongside this segmentation performance, the ThyCLIPNet segmentation stage remains compact, containing 8.55 M parameters and requiring 0.76 GB GPU memory while achieving 78.81 FPS using precomputed BiomedCLIP embeddings (\hyperref[tab:efficiency_analysis]{Table~\ref*{tab:efficiency_analysis}} and \hyperref[tab:reported_efficiency]{Table~\ref*{tab:reported_efficiency}}). Compared with heavier transformer-based methods such as SAMTNU \cite{chen2025samtnu} or DualSwinUNet++ \cite{dialameh2025dualswinunet++}, ThyCLIPNet maintains competitive segmentation performance with a substantially smaller segmentation-stage computational profile. These characteristics support its potential suitability for low-latency ultrasound analysis and resource-constrained computer-aided diagnosis (CAD) pipelines under the adopted BiomedCLIP workflow. In summary, ThyCLIPNet combines biomedical-domain semantic guidance, multi-scale contextual representation, and lightweight attention refinement to provide a favorable balance between segmentation performance and computational efficiency across challenging public thyroid ultrasound datasets. 

\section{Conclusion and Future Work}
\label{sec:conclusion}
In this study, we introduced ThyCLIPNet, a lightweight, semantically guided framework for thyroid ultrasound segmentation that employs a MobileNetV2 backbone for efficient hierarchical feature extraction and selectively integrates BiomedCLIP-derived global image embeddings into the decoder to strengthen semantic consistency while preserving fine structural detail through skip connections. This design is further complemented by ECA-based encoder refinement, an adapted ASPP bottleneck, and two attention variants (custom CBAM and lightweight CBAM) deployed at distinct stages to balance feature refinement and computational efficiency. Across four public thyroid ultrasound datasets (TG3K, TN3K, DDTI, and PKTN), ThyCLIPNet achieves competitive segmentation performance while maintaining a compact segmentation-stage model size and efficient inference under the adopted precomputation workflow. Ablation, repeated-run sensitivity, and efficiency analyses further support the selection of the proposed ThyCLIPNet configuration based on cross-dataset consistency, boundary fidelity (HD95), precision-recall balance, and computational efficiency, rather than isolated improvements on individual datasets or metrics.

Despite these strengths, several limitations remain. ThyCLIPNet currently uses only the BiomedCLIP vision-encoder global image embedding because the evaluated public datasets do not provide aligned radiology reports or structured textual annotations; incorporating unverified external text may introduce semantic noise and institution-specific bias. We also do not employ patch- or token-level representations or cross-attention–based multimodal interactions. While such approaches could improve fine-grained semantic alignment, dense token fusion would substantially increase memory consumption and inference latency, conflicting with the lightweight deployment objective. Finally, although the fixed fusion weight demonstrated stable performance across repeated-run sensitivity and statistical analyses within the evaluated settings, it remains non-adaptive and may offer limited flexibility under highly heterogeneous acquisition conditions or unseen clinical domains. Future work will include cross-dataset and external clinical validation in collaboration with medical experts to further evaluate generalization under unseen acquisition settings and heterogeneous clinical conditions. We also plan to construct clinically verified image-report paired thyroid ultrasound datasets for principled multimodal fine-tuning, investigate lightweight token-level semantic integration strategies, and extend the framework to temporally consistent video ultrasound segmentation under low-latency scanning conditions while preserving the compact and efficient design of ThyCLIPNet.

\section*{CRediT authorship contribution statement}
\textbf{Tasnim Jahan:} Conceptualization, Investigation, Methodology, Resources, Data curation, Software, Visualization, Writing – original draft. \textbf{Md Easin Arafat:} Methodology, Visualization, Validation, Analytical reviewing, Writing – review \& editing. \textbf{Swakkhar Shatabda:} Validation, Writing – review \& editing, Supervision. 

\section*{Declaration of Competing Interest}
The authors declare that they have no known competing financial interests or personal relationships that could have appeared to influence the work reported in this paper.

\section*{Data Availability}
The datasets and models generated during this study are publicly accessible at: \url{https://github.com/Tasnim-Jahan/ThyCLIPNet}

\bibliographystyle{unsrt}
\bibliography{references}

\end{document}